\documentclass[journal]{IEEEtran}

 
 \makeatletter
 \newcommand*\titleheader[1]{\gdef\@titleheader{#1}}
 \AtBeginDocument{%
   \let\st@red@title\@title%
   \def\@title{%
     \bgroup\normalfont\large\centering\@titleheader\par\egroup
     \vskip0.5em\st@red@title}
 }
 \makeatother

\titleheader{\textit{Accepted Version}: IEEE/ASME Transactions on Mechatronics
 \textsuperscript{\textcopyright} 2022 IEEE. Personal use is permitted, but republication/redistribution requires IEEE permission. See \href{http://www.ieee.org/publications_standards/publications/rights/index.html}{\textit{here}} for more information.}
\usepackage[pscoord]{eso-pic}
\newcommand{\placetextbox}[3]{
\setbox0=\hbox{#3}
\AddToShipoutPictureFG{ \put(\LenToUnit{#1\paperwidth},\LenToUnit{#2\paperheight}){\vtop{{\null}\makebox[0pt][c]{#3}}}}
}
\placetextbox{.5}{0.05}{~\copyright 2022 IEEE. Personal use is permitted, but republication/redistribution requires IEEE permission.}
\usepackage[pscoord]{eso-pic}
\placetextbox{.5}{0.03}{{Accepted Version. Paez-Granados \MakeLowercase{\textit{et al.}}}  https://doi.org/10.1109/TMECH.2021.3135453} 

	\usepackage[pdftex]{graphicx}
	\graphicspath{{Figures/}}
    \DeclareGraphicsExtensions{.pdf} 
    
 \usepackage{subfigure}
\usepackage{makecell}

\usepackage{amsmath}
\usepackage{amssymb}
\usepackage{array}

\usepackage{color}

\begin{document}
%

\title{Personal Mobility with Synchronous Trunk-Knee Passive Exoskeleton: Optimizing Human-Robot Energy Transfer}

%
\author{{Diego} {Paez-Granados} $^{1\dagger}$,~\IEEEmembership{Member,~IEEE,}
       Hideki~Kadone$^2$, Modar~Hassan $^3$,
        Yang~Chen $^3$,
        and~Kenji~Suzuki$^4$, ~\IEEEmembership{Member,~IEEE}
\thanks{$^1$ {D.} {Paez-Granados} is with the Swiss Federal Institute of Technology in Lausanne, EPFL, and is a Visiting Researcher at the Artificial Intelligence Laboratory, University of Tsukuba, Japan. {\tt\small dfpg@ieee.org}%
}\thanks{$^2$ H. Kadone is with the Center for Innovative Medicine and Engineering, University of Tsukuba Hospital, Japan.%
       {\tt\small kadone@md.tsukuba.ac.jp}}
\thanks{$^3$ M. Hassan and Y. Chen are with the Artificial Intelligence Laboratory, Department of Intelligent Interaction Technologies, University of Tsukuba, Japan.
        {\tt\small {hassan, chenyang}@ai.iit.tsukuba.ac.jp}}
\thanks{$^4$ K. Suzuki is with the Faculty of Engineering and Center for Cybernics Research, University of Tsukuba, Japan.%
        {\tt\small kenji@ieee.org}}
\thanks{Manuscript received May 31, 2021; revised November 2, 2021; ac- cepted November 23, 2021. Recommended by Technical Editor J. A. Schultz and Senior Editor X. Chen. This work was supported in part by the Ministry of Education, Culture, Sports, Science and Technology of Japan under Grant 2019R300-2 and in part by the Toyota Mobility Foundation under the Mobility Unlimited Challenge grant. (Corresponding author: Diego Paez-Granados.)}
\thanks{This article has supplementary material provided by the authors and color versions of one or more figures available at https://doi.org/10.1109/TMECH.2021.3135453.
Digital Object Identifier 10.1109/TMECH.2021.3135453}
}
%
%
\markboth{IEEE/ASME Transactions on Mechatronics.}%
{Paez-Granados \MakeLowercase{\textit{et al.}}:PERSONAL MOBILITY WITH SYNCHRONOUS TRUNK–KNEE PASSIVE EXOSKELETON}
%
%
%
%
%
\maketitle
%
\begin{abstract}
We present a personal mobility device for lower-body impaired users through a light-weighted exoskeleton on wheels. On its core, a novel passive exoskeleton provides postural transition leveraging natural body postures with support to the trunk on sit-to-stand and stand-to-sit (STS) transitions by a single gas spring as an energy storage unit. We propose a direction-dependent coupling of knees and hip joints through a double-pulley wire system, transferring energy from the torso motion towards balancing the moment load at the knee joint actuator. Herewith, the exoskeleton maximizes energy transfer and the naturalness of the user's movement. We introduce an embodied user interface for hands-free navigation through a torso pressure sensing with minimal trunk rotations, resulting on average $19^{\circ} \pm 13^{\circ}$ on six unimpaired users.
We evaluated the design for STS assistance on 11 unimpaired users observing motions and muscle activity during the transitions. Results comparing assisted and unassisted STS transitions validated a significant reduction (up to $68\%$ $p<0.01$) at the involved muscle groups. Moreover, we showed it feasible through natural torso leaning movements of $+12^{\circ}\pm 6.5^{\circ}$ and $- 13.7^{\circ} \pm 6.1^{\circ}$ for standing and sitting, respectively. Passive postural transition assistance warrants further work on increasing its applicability and broadening the user population.
\end{abstract}
%
\IEEEpeerreviewmaketitle
\section{Introduction}
\IEEEPARstart{P}{ostural} transitions from sit-to-stand and stand-to-sit (STS) are common activities of daily living (ADL) usually overlooked by unimpaired people. However, any impairment or dysfunction on the lower-body results in hindering this crucial activity which initiates locomotion and allows social and physical interaction with the surrounding environment and people. e.g., in cases of motor control dysfunctions, loss of muscle mass (muscular dystrophy), spinal cord injury, or amputation of limbs. All of which require full assistance to recover STS capability.
Moreover, ageing by itself results in making sitting and standing transitions one of the most difficult and energy-expensive ADLs (high metabolic cost \cite{Nakagata2019}) for most of the elderly life \cite{Lindemann2007,Ferrucci2016}. 
Wheelchairs are a mobility solution, however, it completely removes standing from daily life which helps to preserve vital functions of the body, e.g., bone density, bowels function, cardiovascular and respiratory functions. Confirmed in studies of the consequence of sedentary behaviour \cite{Owen2010,Katzmarzyk2012}.

    \begin{figure}[!t]
    \centering
    \includegraphics[width=6.5cm]{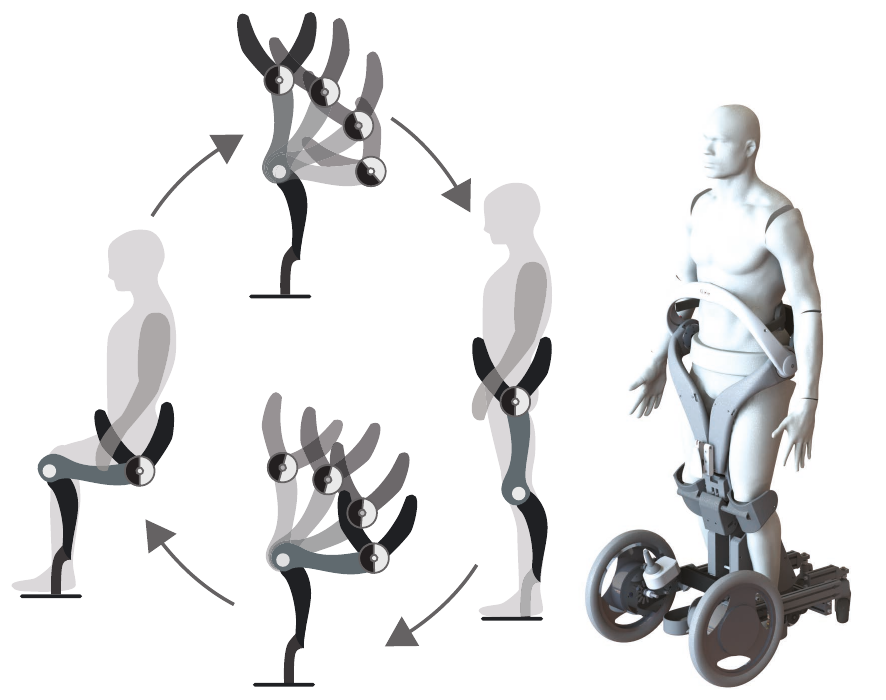}
    \caption{ Synchronous motion of knees joint and pelvis allow a lower-limbs and torso passive exoskeleton for supporting stand-to-sit and sit-to-stand transitions with a single passive element as external energy storage and controlled with natural motions of the upper-body.}
    \label{fig:concept}
    \end{figure}

With this perspective, several assistive devices for STS and personal mobility are currently available such as power standing wheelchairs, and exoskeletons.
Exoskeletons for lower-limb paralysis focus on assisting different mobility impairments, e.g., the robot suit HAL used in rehabilitation scenarios \cite{Tsukahara2015}, the ReWalk exoskeleton for spinal-cord injury (SCI) persons, which relies on crutches for achieving locomotion \cite{Zeilig2012}. While other research approaches envision soft robotic suits for hemiplegia patients \cite{Awad2017} through a flexible exoskeleton for lower-limbs assistance with wire-driven systems \cite{Lee2019,Zhou2018,Zhang2018} and other works as reviewed in \cite{Bao2019}.

Nonetheless, commercially available mobility devices are more common in the form of standing wheelchairs providing readily daily-life solutions that give more autonomy and speed through motorized systems. However, one drawback is the significant power consumption in lifting the user and the heavier the devices become compared with seated powered wheelchairs e.g., TekRMD (Matia Robotics Inc. USA) standing wheelchair weights 120 kg with power autonomy of 5 hours. Therefore, end-users will be driven towards the more standard solution of a seated life.

Previous works have proposed lifting mechanisms for manual wheelchairs through hand-lever propelled systems as in \cite{Shaikh2021}, at the cost of removing standing locomotion.
Our work aims to fill the gap of personal standing mobility through an energy-efficient daily-life mobility device compact and light-weighted, thus giving higher levels of autonomy to a broad range of users.
We propose a mobile base capable of full walking speed and an energy-efficient STS passive exoskeleton with lower-limbs and trunk support by using natural upper-body postures, which is important for the acceptability and daily usage of such a device. 

In previous work, we introduced PAL design \cite{Eguchi2018} a device that assists ankles and knees joints through asymmetric transitions on a two stages STS transition based on static modelling of the human body, also extended to children's usage in \cite{Sasaki2018}.
Nonetheless, support to the torso during the transitions was not present, and it is required for users unable to hold straight postures (torso support) i.e., lack of control over rectus abdominals, erector spinae, or total lack of muscle control on lower-trunk as SCI of level T6 and above. 

In the current work, we extended our previous design in \cite{PaezGranados2018} with a new system that supports the user's torso in both standing and sitting. We propose a novel double-coupling wire-driven parallel mechanism that isolates the sit-to-stand from the stand to-sit synchronized motions.  Herewith, allowing an asymmetric postural transition with torso support (see, Fig. \ref{fig:concept}). Moreover, we present the design for a personal mobility device controlled through a pressure sensing system embedded on the exoskeleton at the abdominal area of the user for hands-free navigation, controlled through the algorithm we introduced in \cite{Chen2020}.

Compared with previously proposed devices for standing locomotion \cite{Eguchi2018,Sasaki2018,PaezGranados2018}, our device coupled pulley system limits the range of users because of the non-linear multi-limb load to the passive element, tested between 55-72 kg (2-gas-springs) and 73-88 kg (3-gas-springs). Whereas the proposed device in \cite{Eguchi2018} supported users from 60 - 100 kg (albeit without torso support). In comparison, standing power wheelchairs such as the LEVO (LEVO Ag, Switzerland) and Tek RMD (Matia Robotics Inc., USA) can lift users up to 120 kg. Nonetheless, our prototype of a passive exoskeleton with active wheels was achieved in a 36 kg compact system, whereas the LEVO is about 170 kg, and Tek RMD is 118 kg device.

The contributions of this paper are mainly two-fold: first, a novel design of a personal mobility device (PMD) for standing locomotion. Consisting of a passive exoskeleton for lower limbs and trunk support on a mobile-powered base controlled by torso motions in a hands-free control system. The novel mechanism is described in detail through a multi-objective optimal design methodology for personalising the human-machine interface modelled through an impedance framework.
Second, we demonstrated the performance of the device through a user study with unimpaired participants analyzing the STS assistance provided by the exoskeleton and compared it with the natural motions of the user.



\section{Method}\label{sec:method}
In this section, we describe the assumptions, objectives, and constraints used for investigating the models of interaction that could achieve the desired assistance in the sit-to-stand/stand-to-sit (STS) postural transitions. 

In the following subsections, we first describe the dynamic model of the human body in the STS and propose the asymmetric postural transition that gives a user control over the motion by natural upper-body postures. 
Subsequently, we formulated the desired assistance for STS postural transition by synchronising the desired motions of the 2 main joints to be supported (namely knees and hip).
Finally, we modeled the mechanism's kinematics and dynamics of the coupled motion with the user to a mathematical description in the context of human-robot interaction including the contact dynamics for simulating the STS transitions herewith, we constructed an optimization framework for evaluating multiple user-body characteristics in the design of the exoskeleton system.

\begin{figure}[!t]
      \centering
		\subfigure[Human biomechanical model (HBM) with marked center of mass to each limb considered for analysis of loads at each joint.]{\includegraphics[width=3.0cm]{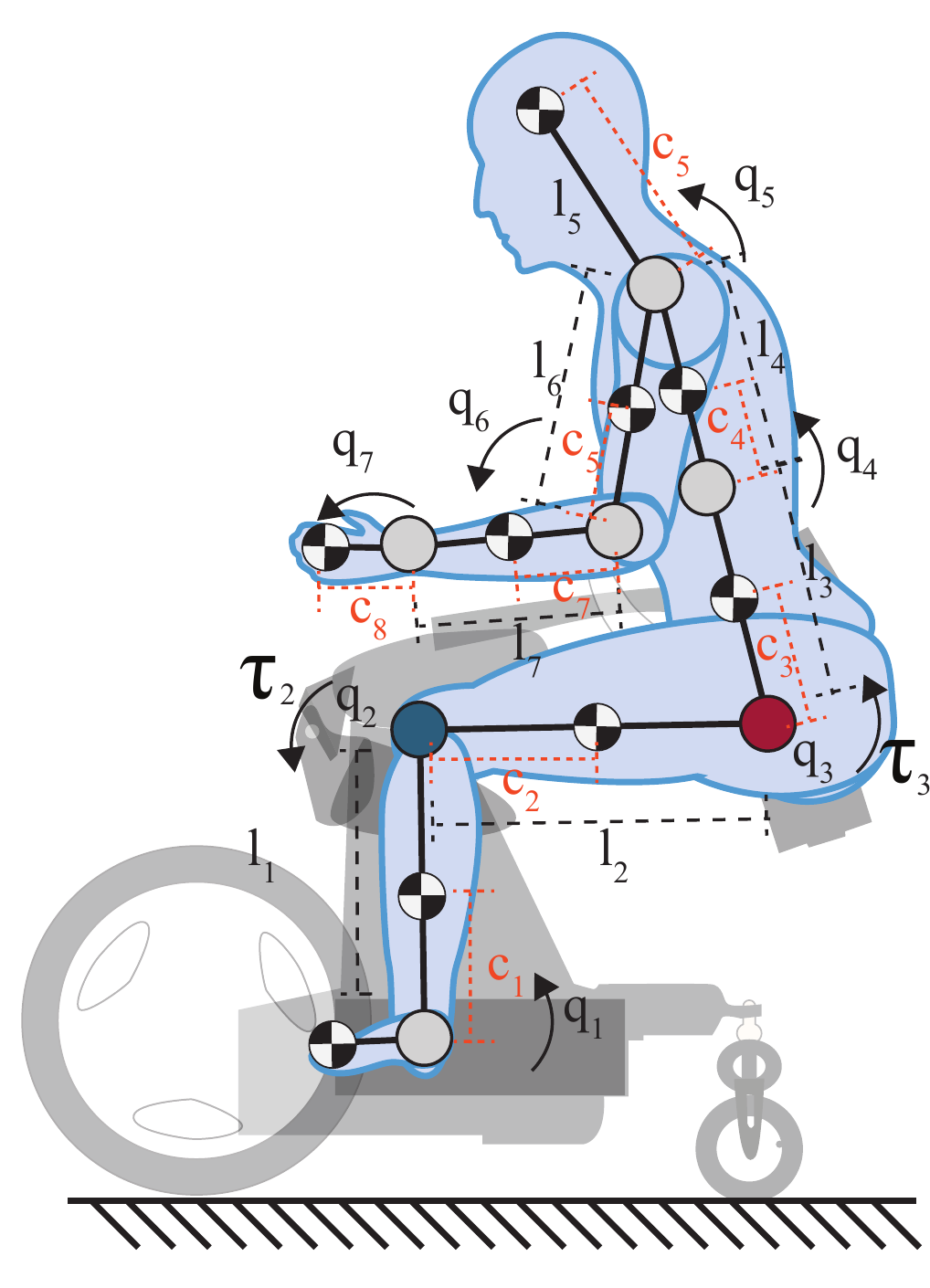}%
		\label{fig:hbm}}
		\hfil
		\subfigure[Natural sitting and standing transitions for 7 users (age 28.7±4.4 years old, height 171.1±6.3 cm, and weight 64± 6.25 Kg). ]{\includegraphics[width=3.4cm]{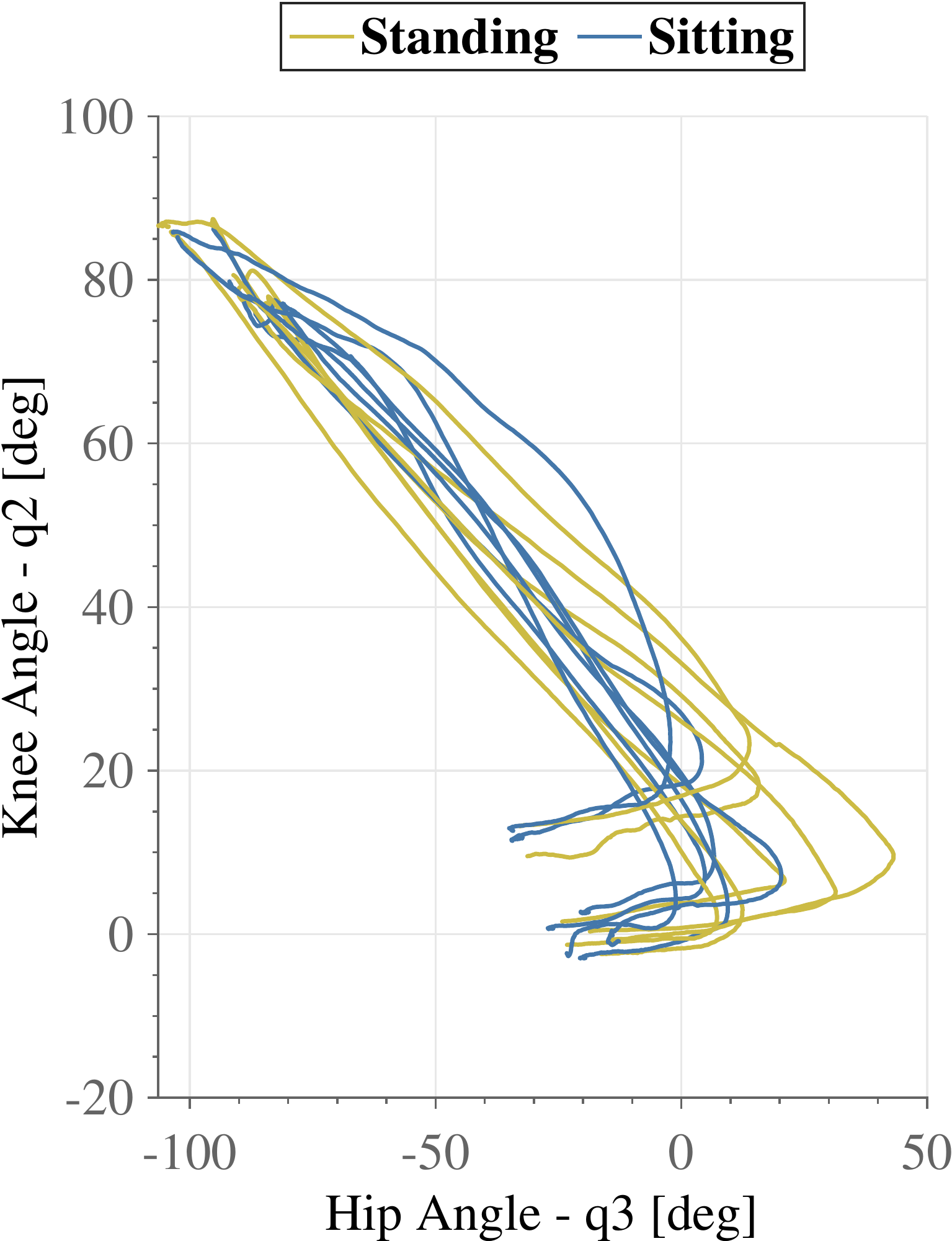}%
		\label{fig:natural_STS}}
      \caption{Proposed simplified human-body model on the Saggital plane during STS transitions for analysis and design of the exoskeleton.}
      \label{fig:model_STS}
   \end{figure}
 
\subsection{Task Dynamics: Human Biomechanical Model}
\label{subsec:task}
Previous works have focused on the joint torque support for STS transitions with 1 DOF torso, minimizing a jerk criterion on a 3-DOF powered system \cite{Asker2017}; or through optimal trajectories of the COM \cite{Geravand2017}, demonstrated with elderly patients in \cite{Werner2020}.
In contrast to previous works where the body model was reduced to 1 degree of freedom (DOF) at the torso \cite{Eguchi2018,PaezGranados2018,Asker2017,Yamasaki2011} we concluded that a 7-DOF human biomechanical model (HBM) was more adequate to fit the transitions shown in Fig. \ref{fig:model_STS}.
Among the target users with spinal cord injuries (SCI) at thoracic nerves T6 and below, cerebral palsy (C.P.) or stroke, many upper-body functions remain feasible and could contribute to the postural transition, therefore, in our work we include crouching of the spine ($q_4$), arms ($q_5,q_6,q_7$) and head in the modelling in order to determine possible postures for controlling the device motion.

We modelled the dynamic load of the user to an exoskeleton supporting this transition through a Lagrangian formulation for an open kinematic chain \cite{Featherstone2008}, as follows:
\begin{eqnarray}
 \tau_{hm} = M_{hm}\left(q_{hm}\right)\ddot{q}_{hm} + \eta_{hm}\left( {\dot{q}}_{hm},{q}_{hm}\right),
\label{eq:Dyn}
\end{eqnarray}
where $\tau_{hm} =[\tau_{1}, \tau_{2}, \tau_{3}, \tau_{4}, \tau_{5}, \tau_{6}, \tau_{7} ]^{T}$ represent the torques in the HBM joint space [ankle, knees, hip, torso, shoulder, elbow, wrist] as $q_{hm} =[q_{1}, q_{2}, q_{3}, q_{4}, q_{5}, q_{6},q_{7}]^{T}$ , as in Fig. \ref{fig:hbm} . With a positive define inertia matrix $M_{hm}$, and the term $\eta_{hm} = C\left(\dot{q}_{hm},q_{hm}\right)\dot{q_{hm}} + g\left( q_{hm}\right) $ accounts for Coriolis-centrifugal, and gravitational forces respectively.
 
The center of mass (COM) for each segment can be estimated by a statistically equivalent serial chain (SESC) as described in \cite{Cotton2009},
\begin{eqnarray}
 \substack{C_{M}\\1} =\frac{m_{1} A^{1}_{0}\lbrace\substack{c_{1}\\1}\rbrace}{M} + \frac{m_{2} A^{2}_{0}\lbrace\substack{c_{2}\\1}\rbrace}{M} +...+ \frac{m_{n} A^{n}_{0}\lbrace\substack{c_{n}\\1}\rbrace}{M} ,
\label{eq:COM}
\end{eqnarray}
where $A^{i}_{0}$ represents an homogeneous transformation matrix from the coordinate system $0$ to $i$, with terms $c_i $ noting the location of the center of mass $m_i$ of the link $i$ w.r.t. the coordinate system $(i-1)$, as depicted in Fig. \ref{fig:hbm}. And the term M denotes the total body mass as $M = \sum^{n}_{0} m_{i}$, for $i \in R^{n}$.

We recorded the natural STS postural transitions of 7 unimpaired participants showing in Fig. \ref{fig:natural_STS} and extracted the loads and synchronous motion of the knee and hip using the proposed model.
The main takeaway from these data was a common leaning forward method in the sit-to-stand transition where the hip angle - $q_3$ - reached angles between 30 to 45 degrees; validating a natural method of postural transitions through leaning the upper body for minimizing the knee joint torque.

\subsection{Synchronous Asymmetric Postural Transitions} \label{ss:postural_transtion}
 \begin{figure}[!t]
 	\centering
		\includegraphics[width=8.8cm]{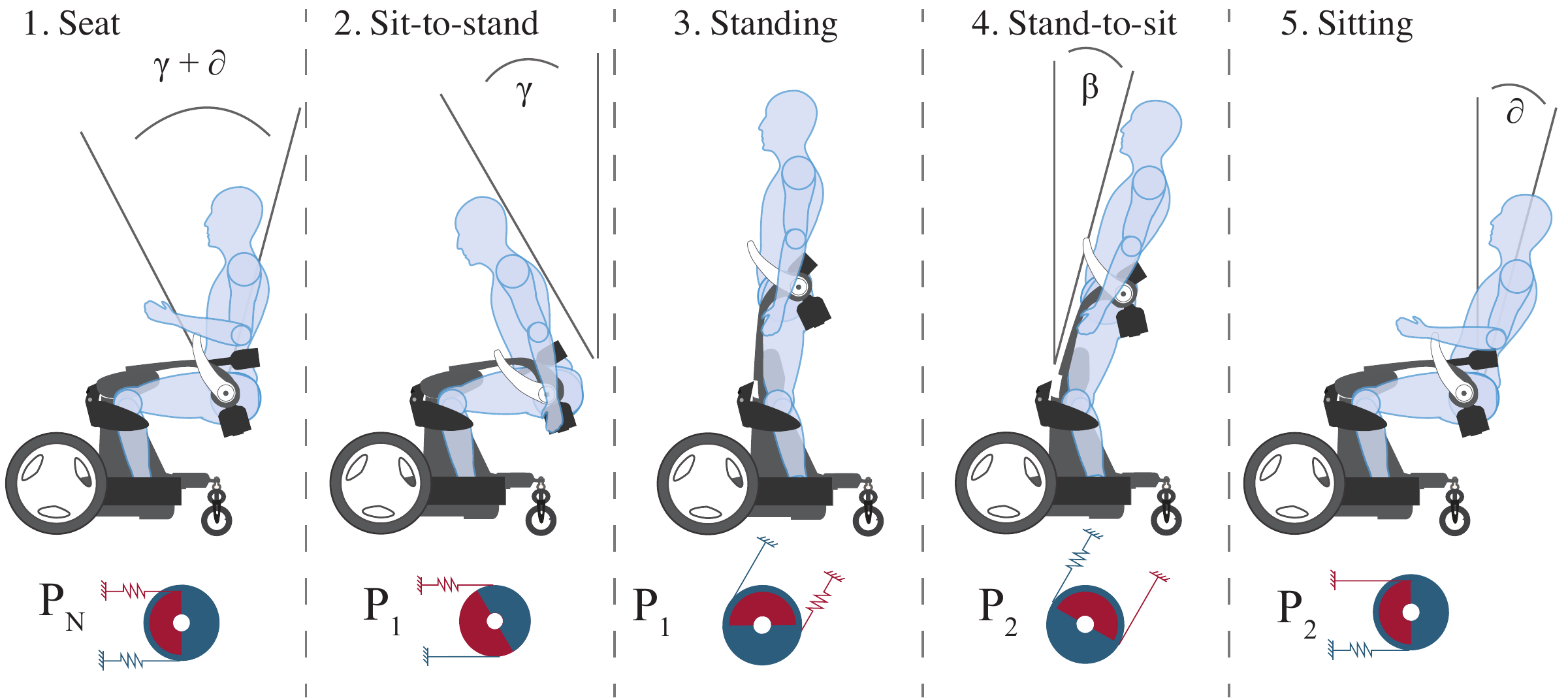}
  		\caption{Postural transition support from sit-to-stand. State 1: the seated natural state where a user can freely move their torso, neither pulley system is engaged ($P_N$). State 2, sit-to-stand transition engaging through a torso's forward inclination ($\gamma$) enabling the first pulley system ($P_1$) in blue for torso support. State 3, standing. State 4, transition stand-to-sit with through a leaning backward ($\beta$) required for engaging the second pulley system ($P_2$). Finally, state 5 is the sitting posture showing the final torso angle ($\delta$). Afterwards the cycle restart to state 1.
	\label{fig:transition}}
   \end{figure}
In this work, we proposed to achieve the STS transitions on asymmetric standing (leaning forward) and sitting motions (leaning backward), which resulted from targeting the following objectives:
	\begin{enumerate}
			\item Standing and sitting support for lower-limbs and trunk. 
			\item Natural transition: controlled by the upper-body motion.
			\item Synchronous motion of lower-body and upper-body.
			\item Comfortable transitions: minimal jerk and max fit of actuator profile to load.
			\item Low-weighted device: smallest possible passive element at a single location.
	\end{enumerate}
Figure \ref{fig:transition} depicts the result of our design for passive transitions. \textbf{Stage 1}, gives seated freedom to the user for moving his/her torso angles $\gamma + \delta$, with a safety lock on the exoskeleton to avoid undesired motions.
\textbf{Stage 2}, starts by releasing the mechanical lock on the exoskeleton, followed by a leaning forward of the torso of an angle $\gamma$ driven by the user. Afterwards, the movement of the knee joint should be coupled to the hips (represented by the blue pulley) making the exoskeleton's motion synchronous during the transition.
\textbf{Stage 3} at the standing posture, the exoskeleton reaches a mechanical limit and the motion is halted. The user has leaning backwards freedom $\beta$, and the same mechanical lock to avoid undesired transitions from the exoskeleton.
\textbf{Stage 4}, the user unlocks the exoskeleton and leans backwards engaging at an angle $\beta$ where the system couples the hips to knee joint motions (depicted in red in Fig. \ref{fig:transition}), herewith giving support to the back of the user.
\textbf{Stage 5}, this is the final state where the user arrives at sitting with an angle $\delta$ on their torso.

\subsection{Coupling-Decoupling Wire-driven Parallel Joint }\label{ss:mechanism}
Preliminary testing with spinal cord injury volunteers on our previous prototype \cite{PaezGranados2018} showed that several users required support at their torso on both standing and sitting transitions because of the lack of control over their abdominal and back muscles. This varies significantly by the level of injury from T10 to T6 and each patient unique remaining muscle control.

\begin{figure}[!t]
 	\centering
		\includegraphics[width=6.5cm]{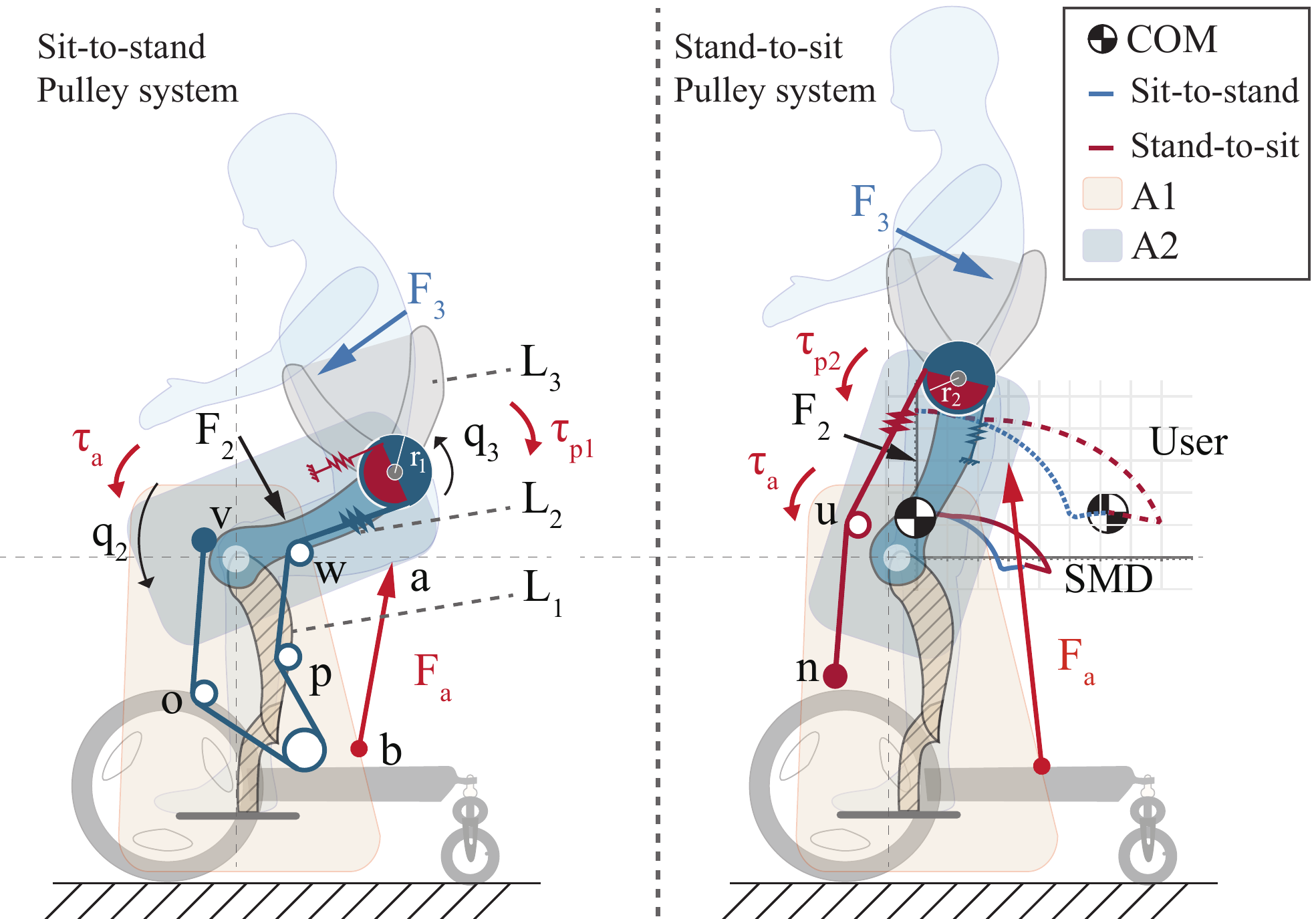}
  		\caption{Exoskeleton's optimization variables: we propose a double wire pulley mechanism that couples the motions at $q_2$ (knees) and $q_3$ (hips). The blue pulley system $P_1$ for sit-to-stand (left-side), couples the hips joint $q_3$ to $q_2$, and controls the torque transfer through ${|p-w|}$, and ${|v-o|}$. Transferring load from link $L_2$ ($A_2$) to the base link $L_1$ ($A_1$). Correspondingly, the red pulley system $P_2$ (right-side), drives $q_3$ coupled to $q_2$ by connecting ${|n-u|}$, for the stand-to-sit transition. The whole COM expected movement in both transitions for the PMD w/ user and the user alone is depicted on the right side.
	\label{fig:exo_model}}
\end{figure}

Therefore, we propose a double wire-driven parallel mechanism to provide torso support in both transitions through the asymmetric motion where hips to knee coupled motions of sitting and standing differ in the expected coupled angles at the hips [$\gamma \sim -\pi/2$] for standing, and [$-\pi/2+\beta \sim \delta$] for sitting.
We found wire-driven transmissions a light-weighted solution to transfer power and couple multi-limb motions, as used in multi-joint serial robots, such as the WAM (Barret Technology, USA) and some novel joint transmission systems \cite{Jiang2018}.
Here, the proposed design creates a controlled coupling-decoupling motion of two joints, making it directional dependent through a set of two different pulley circuits with independent length constraints that achieve the desired coupling angles, as illustrated in Fig. \ref{fig:exo_model}. Each transmission system engages separately for standing and sitting through the blue wire-pulley system $P_1$  and the red system $P_2$, respectively.

The first pulley system $P_1$ (marked on a blue path for standing torso support in Fig. \ref{fig:exo_model} left-side) operates by anchoring at a pulley on the hips joint $q_3$ (radius $r_1$) through a wire circuit exiting link 2 at $w$ and connecting to the base link 1 at $p$, the wire exits link 1 at $o$, and anchors at $v$ on link 2.
The change in distance $\Delta (|p-w|+|v-o|)$ between sitting ($q_2=q_o$) and standing ($q_2=q_f$) determines the output motion at $q_3$, from $\gamma \sim q_s$.
This closed pulley circuit allows to partially balance the moment non-linearity at $q_2$ caused by having the coupled motion of both joints.

Then, the leaning forward angle ($\gamma$) for engaging the first wire-pulley allows the motion to be driven by:
	\begin{eqnarray}
		q_{3}\vert_{\gamma}^{q_{s}} = \alpha_1  q_{2} \vert_{q_{o}}^{q_{f}}, \label{eq:stand_q3_q2}
	\end{eqnarray}
where $\alpha_1$ represents a linear gain, $q_s$ denotes the final hip joint in standing, $q_o$ and $q_f$ represent the knee angle at sitting and standing. 

Similarly, the second pulley system $P_2$ (marked on a red path for sitting torso support in Fig. \ref{fig:exo_model} right-side) couples the motion through a pulley at $q_3$ (radius $r_2$) and the wire exiting link 2 at $u$, and fixing at $n$ on link 1. Thus, the distance change $\Delta |n-u|$ between standing ($q_2=q_f$) to sitting ($q_2=q_o$) determines the torso motion $q_3$, from $\beta \sim \delta$.

Then, the leaning backwards an angle ($\beta$) leads to the engaging the second mechanism whereas the first wire-pulley system remains out of range, thus the motion is driven by a second system:
	\begin{eqnarray}
		q_{3}\vert_{\beta}^{\delta} = \alpha_2  q_{2} \vert_{q_{f}}^{q_{o}}, \label{eq:sit_q3_q2}
	\end{eqnarray}
where $\alpha_2$ denotes an independent linear gain, and $\delta$ denotes the final hip joint at sitting ($q_2 = q_o$). 

\begin{figure}[!t]
 	\centering
		\includegraphics[width=8.0cm]{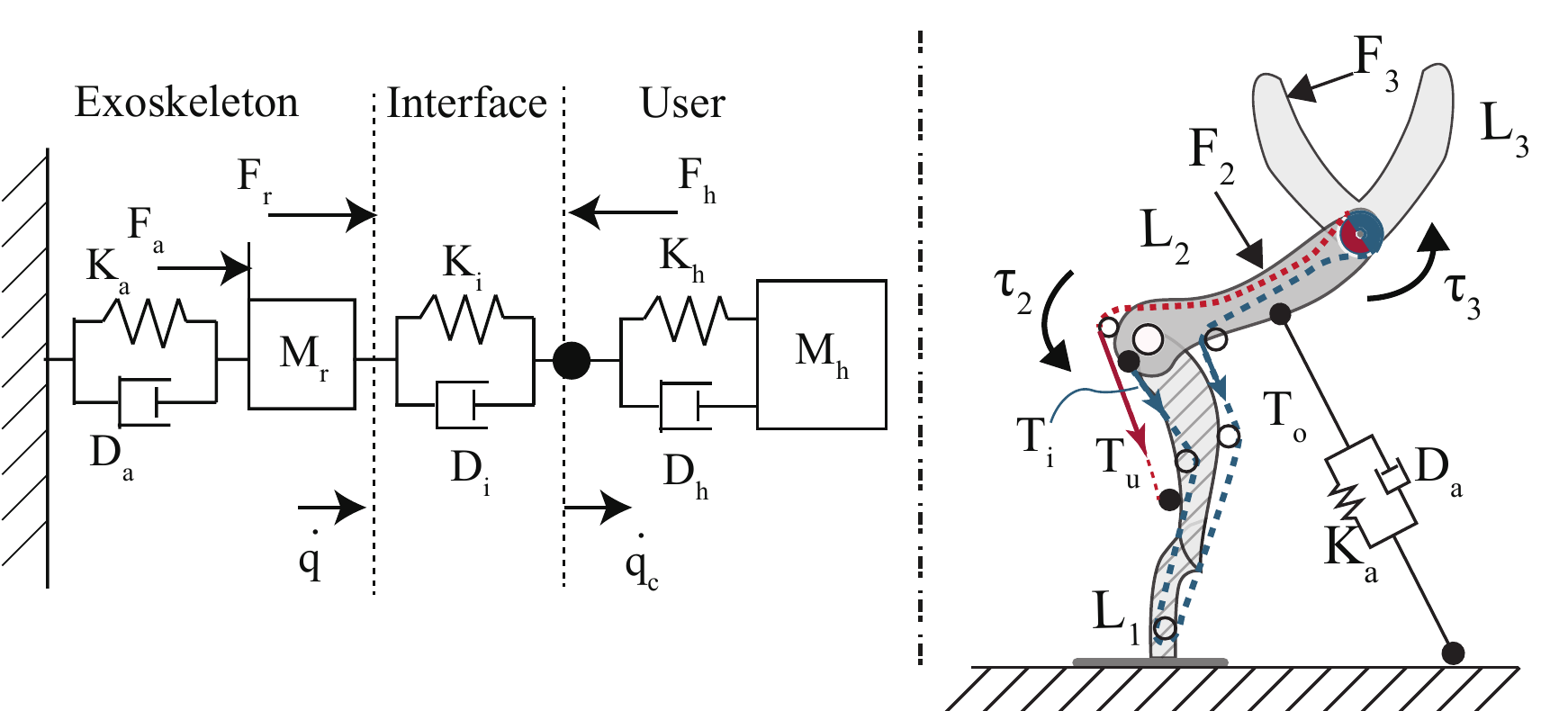}
  		\caption{Left side: Abstraction of the exoskeleton interaction with a user through an impedance model. The user load $F_h$, represents the human as $M_h$, $K_h$, and $D_h$ interfacing with a coupling $K_i$, $D_i$ which accounts for the mechanical compliance of the structure and link. The exoskeleton $M_r$ is controlled by an actuator $F_a$ (resultant of $K_a$, $D_a$). Right side: free body diagram of the moment loads at the knee joint $q_2$, showing the standing pulley system forces (blue), and sitting pulley system forces (red).
	\label{fig:impedance_model}}
\end{figure}

The location of these wire-pulley systems allows us to control the observed momentum at the knee joint, obtained as the sum of moments at $q_2$:
	\begin{eqnarray}
		M_{o}\vert_{q_{s}}^{q_{f}} = -\tau_{2} + (v\times \bar{T_{i}}) + (w\times \bar{T_{o}}) 
		, \label{eq:load_standing}
	\end{eqnarray}
where $v $ and $w$ represent the location of the pulley system input and output on link 2. 
$T_{o}$ denotes the force been transferred in the direction $\vert v-o \vert $ and $T_{i}$ the output force in the direction $\vert p-w \vert$, as depicted in Fig. \ref{fig:impedance_model} right-side.
The sitting wire-pulley system (depicted in red) transfers upper-body load ($\tau_3$) to the knee joint through the connection in $T_u$ (see, Fig. \ref{fig:impedance_model}), thus, the load at the knee joint can be written as:
\begin{eqnarray}
		M_{o} \vert_{q_{f}}^{q_{s}}= -\tau_{2} + (u\times \bar{T_{u}})
		, 		\label{eq:load_sit}
	\end{eqnarray}
where $u$ represents the location of the output pulley on link 2, and $T_{u}$ the output force in the direction $\vert n-u \vert$ (shown in Fig. \ref{fig:exo_model}).
    
\subsection{Human-Robot Interaction Multi-objective Optimization} \label{subsec:HR_optm}
The objectives proposed in section \ref{ss:postural_transtion} have competing trade-offs of linearity, power transfer and user motion requirement through the proposed double-wire mechanism. 
Therefore, we modeled the STS transition considering the dynamics of the interaction and created a framework for simulating towards an optimal solution for both sitting and standing assistance mechanisms.

We set the features in $\xi = [u,v,w,n,o,p,r_1,r_2,\eta]$ as minimization arguments accounting for transmission through the standing wire pulley systems $P_1$ ($v,o,p,w,r_1$), the sitting wire pulley systems $P_2$ ($n,u,r_2$), and the efficiency of the transmission $\eta$.
We optimize through the location of the vectors$\vert p-w \vert$,  $\vert v-o \vert $ and the force-torque transferring ratio at the hip joint $r_1$ for standing.
While $\vert u-n \vert$ and the transmission ratio $r_2$, for sitting.
\textbf{\textit{Constraints:}}
Location constraints were applied to achieve a feasible solutions around the user's body limiting the size of the exoskeleton. 
$(u,v,w) \in \mathbb{R}^2$ were constrained to an area $A_2$ ( $u \wedge v \wedge w \in A_2 $). 
$(n,o,p) \in \mathbb{R}^2$ correspond to the base link $l_1 $ constrained to $A_1$ ($n \wedge o \wedge p \in A_1 $).

As well, an equality constraint was set to enforce the start to end synchronous motion between of $q_2$ and $q_3$, for sitting and standing through: 
	\begin{eqnarray}
		\Delta (\vert p-w \vert - \vert v-o \vert ) \vert_{q_{o}}^{q_{f}}=r_1  \Delta q_{3std}\vert_{\gamma}^{q_{s}}, \label{eq:length1}\\
        \Delta (\vert n-u \vert ) \vert_{q_{f}}^{q_{o}} =r_2  \Delta q_{3sit}\vert_{\beta}^{\delta}, \label{eq:length2}
	\end{eqnarray}
where $r_1$ represents the radius of the output pulley at the hip joint $q_3$, and $\Delta q_{3std} = \gamma - q_s$ denotes difference between engaging angle at the leaning forward motion $\gamma$ and the final standing posture angle $q_s$ for standing transition, while $\Delta q_{3sit} = \beta - \delta$ denotes sitting transition difference from the leaning backward angle $\beta$ and the sitting pose $\delta$.

The optimization framework for achieving the best solutions within the space of feasible arrangements for the exoskeleton was evaluated by a Pareto-optimal front explored through a many-objective non-dominated sorting genetic algorithm (NSGA-II) \cite{Deb2002}. An exploration algorithm validated in convergence towards the actual optimal Pareto front in design examples, even under multiple state scenarios \cite{Deb2018}.
This algorithm allowed us to mutate over the possible populations of solutions through a definition of the objectives and associated costs for achieving the proposed mechanism, evaluated through the forward dynamics of the system as modelled in the previous section.

We model the interface through an impedance interaction at the two joints in contact with a damped interface, as it has been applied for stable physical human-robot interaction in whole body impedance control \cite{PaezGranados2017}, and introduced in our previous work \cite{PaezGranados2018}.
\begin{equation}
		\tau_{hr} = M_{h}\ddot{q_c} + D_{h}\dot{q_c} + K_{h}q_c + \eta_{i}\left( \dot{q_c},{q_c}\right), \label{eq:imp_human}
	\end{equation}
    where the $\eta_{i}$ represents the non-linear terms defined in (\ref{eq:Dyn}) for joint $i \in R^n$, considered for the serial kinematic chain.  Nonetheless, for a user with paralyzed limbs the stiffness $K_h$ and damping $D_h$ of each joint are insufficient for holding the upright posture, thus, compared with the exoskeleton's it could be neglected yielding:  $\tau_{hr} = M_{h}\ddot{q_c} + \eta_{i}\left( \dot{q_c},{q_c}\right)$. 
Herewith, the equation of motion for the exoskeleton with a torque input from the actuator to the human body can be expressed as:
	 \begin{equation}	
		\tau_{hr} = (M_h+M_r)\ddot{q} + D_i\dot{q} + K_i q + \eta_{i}\left( \dot{q},{q}\right),
		\label{eq:imp_robot}
	\end{equation}
where $\tau_{2} = \tau_a$. And the effective torque at the joint is defined by the location of the linear actuator $a$ and $b$, as depicted in Fig. \ref{fig:exo_model}. Thus, $\tau_a =  \bar{a} \times \bar{F_a}$, where the actuator's force is defined ideally by the linear equation:

\begin{equation}
		F_a = (f_0 +  k_a  \Delta x) \eta_t + D_a \dot{x} , \label{eq:Fa}
	\end{equation}
where $D_a$, and $k_a$ denote the spring-damper characteristics of the actuator, $f_0$ represents the zero crossing force, and $\eta_{t}$ represents the whole power transfer efficiency of the wire-driven system. 

Using the above model for the moment load (\ref{eq:imp_robot}), transmission system (\ref{eq:load_standing} and \ref{eq:load_sit}) and actuator profile (\ref{eq:Fa}), we introduce the following definitions of the objectives for optimization:

\subsubsection{Objective 1 - Minimal moment load at the knee joint}
We defined the overall effective moment load ($M_o$) at the knee joint - $q_2$ in terms of the design parameters encapsulating the locations of each connection point between the thighs link ($L_2$) and the base ($L_1$), for both standing and sitting transitions as:
		\begin{equation}
		\begin{split}
        \substack{{\mathrm{arg min}} \\{\xi} } \quad   \frac{1}{M_{r} m} \sum_{q_2=q_{o}}^{q_{f}} M_o(\xi,\tau_{hr}, q_2), \\
        \substack{{\mathrm{arg min}} \\{\xi} } \quad \frac{M_{r}}{m} \sum_{q_2=q_{f}}^{q_{o}} \frac{1}{M_o(\xi,\tau_{hr}), q_2}, 	
        \label{eq:obj_Mload}
        \end{split}
	\end{equation}
where $q_o$ denotes the sitting angle and $q_f$ the standing angle at the knee joint, 
$M_o$ represents the moment load at the knee joint of the exoskeleton,
$M_r$ corresponds to a normalization constant (max load for a 90 kg user),
and $m$ corresponds to the number of samples used in the discrete space of $q_2 = [q_o \sim q_f]$, as $m = (q_f-q_o)/dq$ for a sample rate $dq$.

\subsubsection{Objective 2: Natural torso motion}
We minimized the difference to the natural posture transition based on the captured data presented in Fig. \ref{fig:hbm}, which we approximated as a linear motion of the knees ($q_2$) and hips ($q_3$) within the range $q_3 =[ \gamma \sim q_{s}]$ for standing, and $q_3 = [ \beta \sim \delta ]$ for sitting. Therefore, we minimize a linear fitting as follows:
    \begin{equation}
        \begin{split}
            \substack{{\mathrm{arg min}} \\{\xi} }\quad \frac{1}{ (\gamma - q_s) m} \sum_{q_2=q_{o}}^{q_{f}} \Vert q_3(\xi, q_2) - q_{l}(\gamma,q_{s},q_2) \Vert, \\ \substack{{\mathrm{arg min}} \\{\xi} }\quad \frac{1}{ ( q_s - \beta) m} \sum_{q_2=q_{f}}^{q_{o}} \Vert q_3(\xi, q_2) - q_{l}(\beta,\delta,q_2) \Vert,
		 \label{eq:obj_q3}
		 \end{split}
	\end{equation}
where $q_{3}$ denotes the coupled hip joint during the transitions, $q_{s}$ the final standing angle.

\subsubsection{Objective 3: Matching a linear torque profile}
We define both standing and sitting as,
    \begin{equation}
    \begin{split}
        \substack{{\mathrm{arg min}} \\{\xi,q_{hr}} }\quad \frac{1}{ M_{n} m} \sum_{q_2 = q_{s}}^{q_{f}} \Vert M_o(\xi,\tau_{hm},q_2) - M_l(q_{o},q_{f}) \Vert,  \\
		\substack{{\mathrm{arg min}} \\{\xi,q_{hr}} }\quad \frac{1}{ M_{n} m}\sum_{q_2 = q_{f}}^{q_{s}} \Vert M_o(\xi,\tau_{hm},q_2) - M_l(q_{f},q_{o}) \Vert,
		 \label{eq:obj_lin}
	 \end{split}
	\end{equation}

where $M_l(q_{o},q_{f})$ denotes the ideal moment at the knee joint for a passive linear actuator (the gas-spring) with force equation in (\ref{eq:Fa}).

\begin{figure}[!t]
    \centering
    \includegraphics[width=7.0cm]{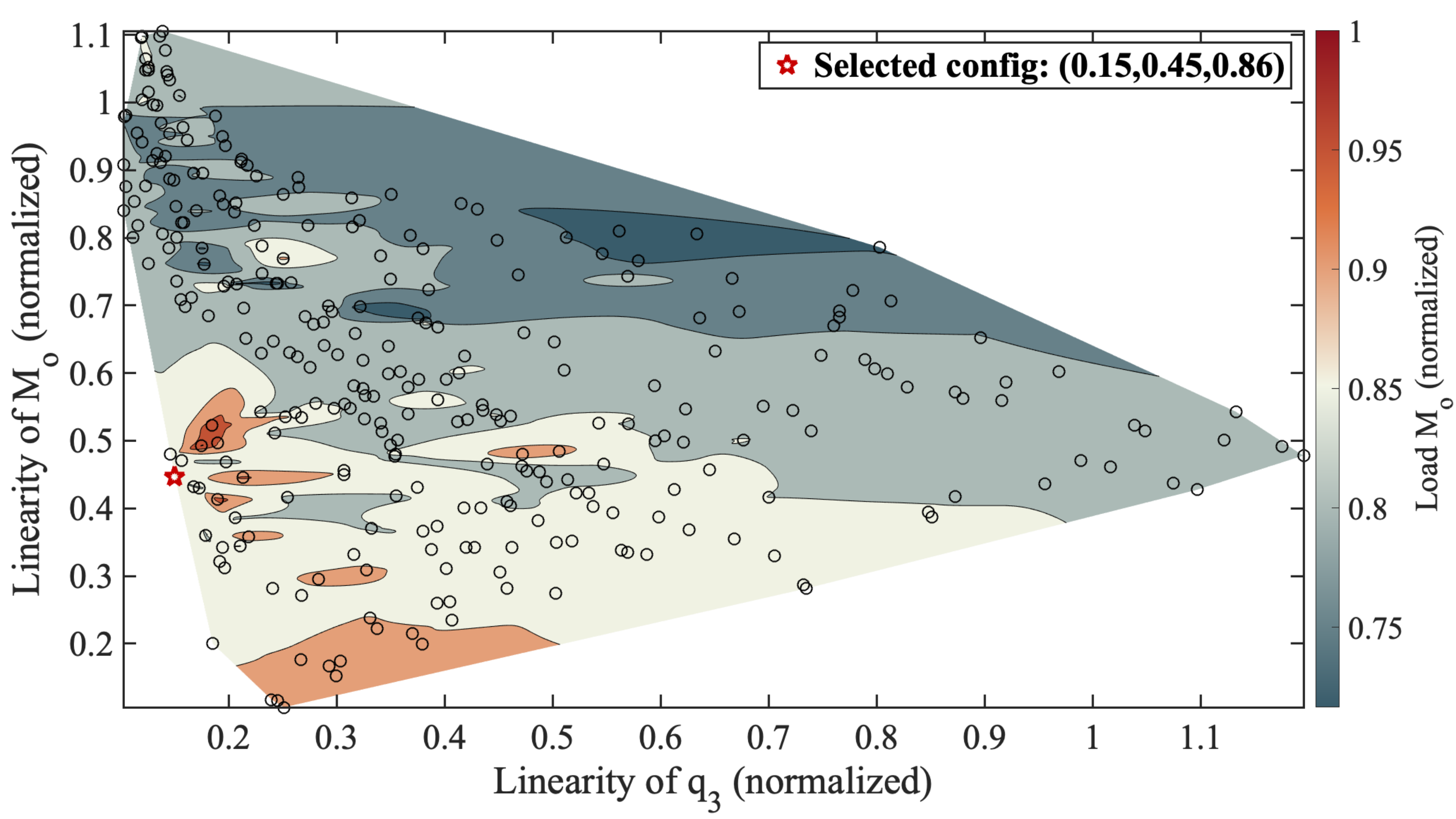}
    \caption{Pareto-optimal front of possible configurations to achieve the synchronous transitions with the proposed objectives. The black dots represent the discrete solutions for each objective: moment load at $q_2$ (\ref{eq:obj_Mload}), torque linearity at $q_2$ (\ref{eq:obj_lin}), and motion linearity between knees and hip (\ref{eq:obj_q3}). The surrounding contour was derived by scatter interpolation. The chosen solution was set at the marked red star, where we chose based on the linearity of the motion and a mid-point to other objectives.
    }
    \label{fig:pareto}
\end{figure}
The results in Fig. \ref{fig:pareto} present a set of contours that map the Pareto front with black circles for the three objectives.
The trade-off on objectives 1, 2, and 3 exists as a result of the task dynamics and parallel mechanism, where the motion control could be achieved by either reducing significantly the moment load at the knee joint or the motion linearity, and the moment linearity is affected by the distribution of the pulley system. 
If a single objective were optimized, e.g., the moment load (as in our previous prototype \cite{PaezGranados2018}) the linearity of the hip-to-knees joint motion would be drastically reduced in the transition.
\subsubsection{Minimal Actuator Force}
Finally, the location of the passive actuator is independently optimized to maximize the torque output:
\begin{eqnarray}
 		\substack{{\mathrm{arg max}} \\{\Phi} } \quad  \sum_{q_2=q_{o}}^{q_{f}} \tau_a(\Phi),	\label{eq:obj_Fa}
	\end{eqnarray}
	where $\Phi = [F_a, \bar{a}, \bar{b}, n]$
$\bar{ba}$ was chosen so that the actuators effective force $F_a$ would satisfy the required torque to allow both sitting and standing transitions.
Therefore the constraints were set for standing load $M_{o}(q) \mid^{q_2=q_o}_{q_f}$ and sitting load $M_{o}(q)_{q_2=q_f, q_o} $ as,
	\begin{eqnarray}
		\tau_{a} > M_{o}(\xi, q_{hr}) \mid^{q_2=q_o}_{q_f}, 		\label{eq:constr1}\\
		\tau_{a} < M_{o}(\xi, q_{hr})\mid^{q_2=q_f}_{q_o}	\label{eq:constr2}
	\end{eqnarray}
Moreover, the actuator was selected from existing vendors database, thus, discretely selected to fit in the constrained area $A_2$ and $A_1$. 

\subsection{User Embodied Control Interface} \label{ss:embodied}
\begin{figure}[!t]
\centering
\includegraphics[width=7.0cm]{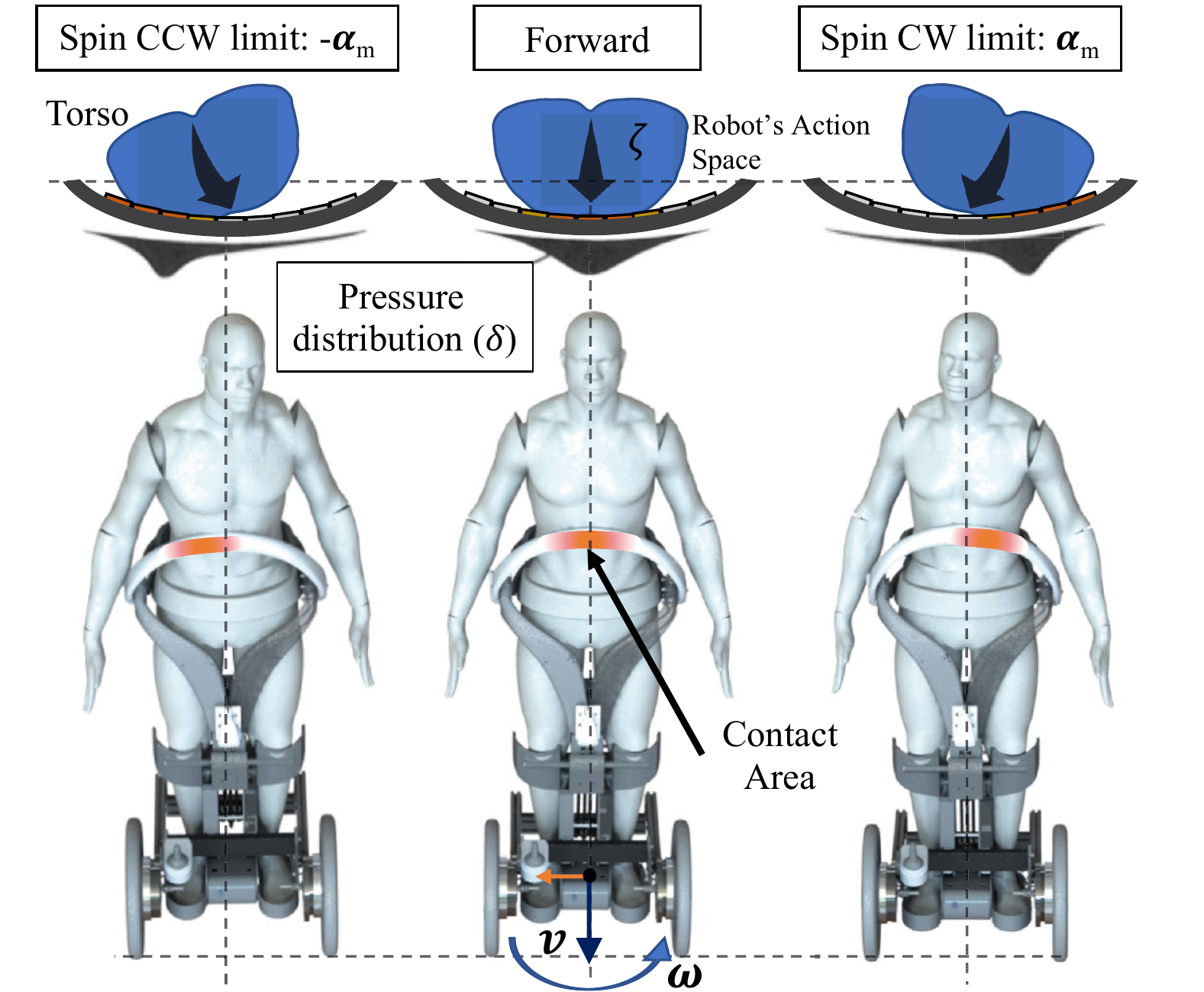}
\caption{User control interface by a pressure sensing of upper-body motions which maps natural body postures to low-level motion control of the in-wheel motor system for driving mobility device.}
\label{fig:embodied_motion}
\end{figure}

The exoskeleton system was designed as a personal mobility device (PMD) with a user intention recognition embedded on the torso support bar (link 3) through pressure sensing system. 
We propose a hands-free navigation method with small upper body motions by continuous mapping of the user's torso to the mobile exoskeleton's local coordinate system at the wheels (as shown in Fig. \ref{fig:embodied_motion}) by following the principle of gaze tracking during walking \cite{Hollands2002}, which showed that a person would normally follow their walking direction with their head and torso; matching a natural walking motion while exploiting the user's upper-body residual capabilities.
We mapped the upper-torso motion counter-clockwise (CCW) to a rotation CCW, a spinning clockwise (CW) to a CW rotation, and slight forward leaning of the torso to a forward driving at the base of the robot. 
The backward motion was defined by a conscious independent action of the hands with the use of the sensors at each extreme of the torso bar by pressing them simultaneously, thus, avoiding unintended activation.

As depicted in Fig. \ref{fig:embodied_motion}, we constructed the inner surface of the support bar with a soft material and an array of pressure sensors for mapping the body postures to a locomotion intention. 
The intention recognition algorithm maps the 2D projection of the pressure distribution changes to the robot's control space in velocity $\zeta = [v, \omega]'$, where $v$ denotes the linear, and $\omega$ the angular velocity, both in the local frame of the robot.
We defined the input space for the user $u=[ \rho, P]$ as the center of pressure ($ \rho$) and the maximum pressure ($P$) over the sensors' array. 
Subsequently, we used a proportional control to drive the robot $ \lVert \zeta \rVert$ with the user given pressure $P$, $\lVert \zeta \rVert = K P$, with a gain matrix $K = [k1, k2]'$, constrained to the velocity limits of the mobility device ($v_{max},  \omega_{max}$). A detailed description of the sensor mapping $v(\rho), \omega(\rho)$ and its generalization used in this controller can be found in our previous work \cite{Chen2020}.

\section{System Configuration: Personal Standing Mobility }\label{sec:system}

\begin{figure}[!t]
      \centering
  		\subfigure[Moment load of the user at the knee joint ($q_2$) compared to the knee standing joint angle.]{\includegraphics[width=6.5cm]{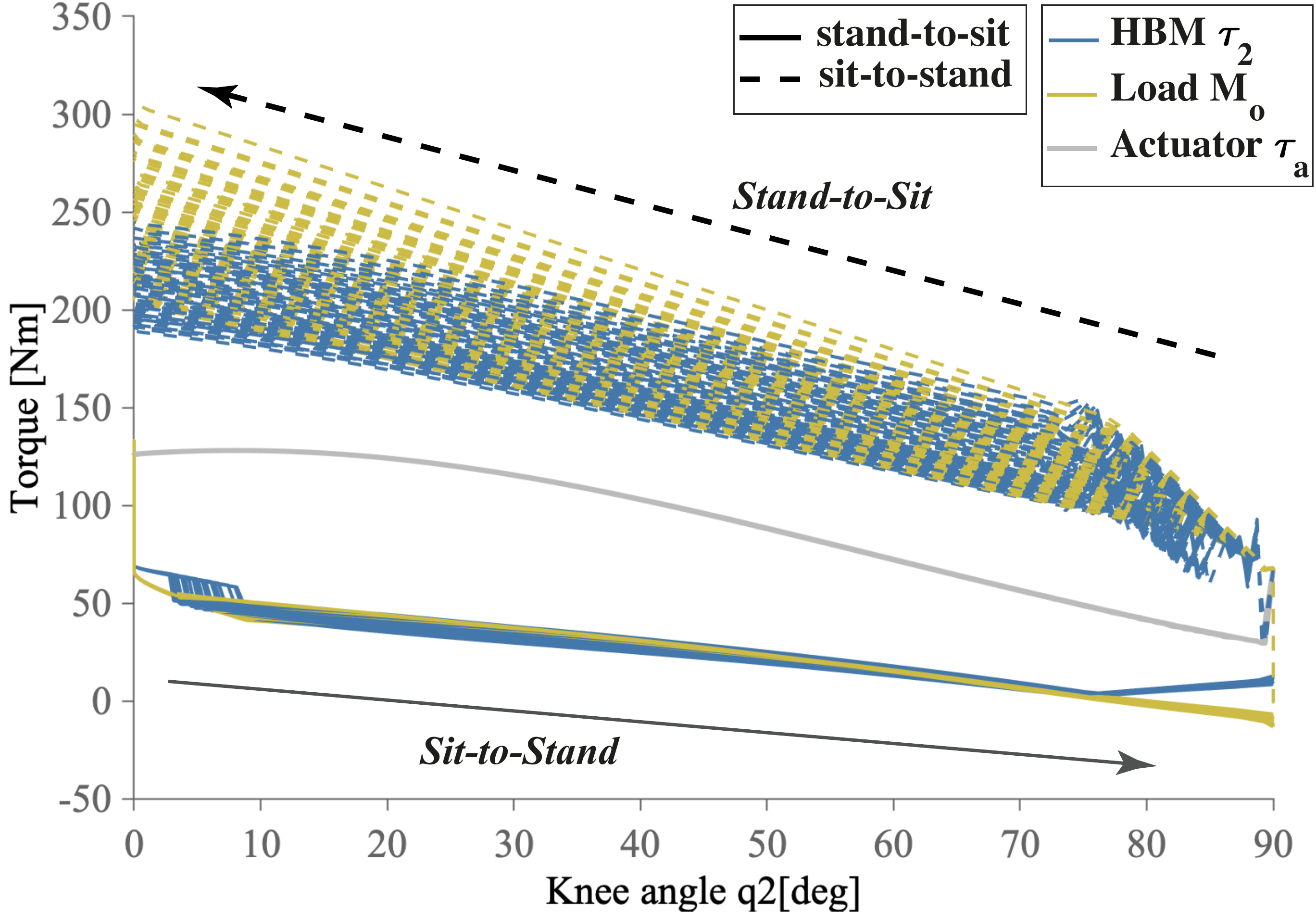}%
		\label{fig:load_q2}}
		\hfil
		\subfigure[Moment load at the knee joint ($\tau_2$) during the stages of STS as depicted in Fig. \ref{fig:transition} for multiple user's weight and height. ]{\includegraphics[width=6.5cm]{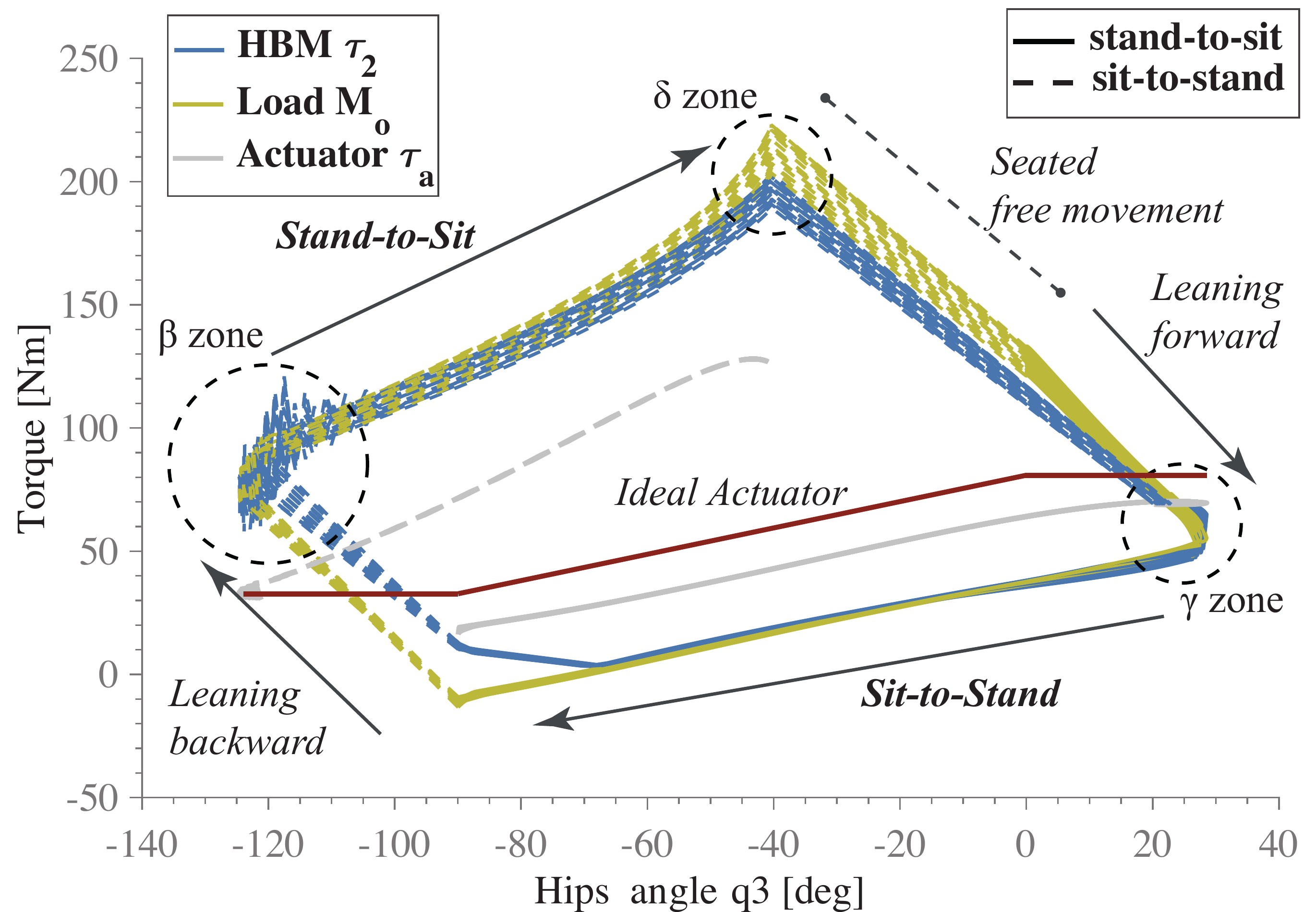}%
    		\label{fig:load_q3}}
		\hfil
		\subfigure[COM motion required for operating the passive exoskeleton during STS transitions for users $42 - 62$ kg with height range $1.40 - 1.80$ cm.]{\includegraphics[width=6.0cm]{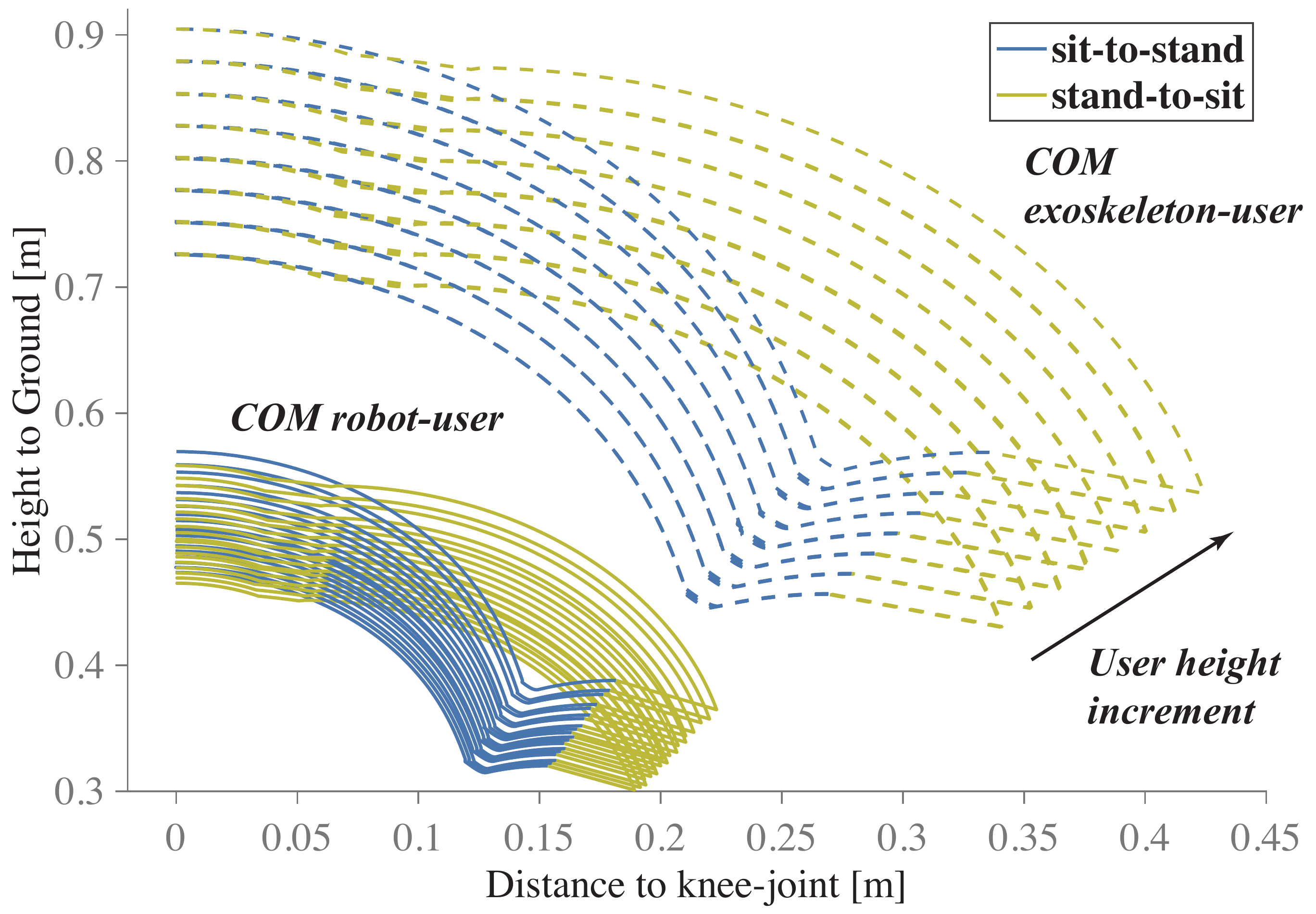}s%
		\label{fig:COM}}
      \caption{Pareto-optimal implementation and evaluation of potential user usability gap through the forward dynamic model and estimated weight/height distributions.}
      \label{fig:sts_simulation}
   \end{figure}

Results from the selected configuration of pulley systems and actuators are shown in Fig. \ref{fig:sts_simulation}. We chose a value with higher priority to the motion linearity with the compactness of the location of the wire-driven system. One of the lowest moment load to linearity configurations was chosen (the red mark in Fig. \ref{fig:pareto}).
The simulation of multiple users' STS transitions with the user load to the knees in yellow ($M_o$), and the exoskeleton with user transfer power to the knee joint plotted in blue ($\tau_2$).
The motion from the sitting posture starts on the top of the figure with loads over 200 Nm, which are controlled by the user through leaning backwards or forward. If the user-controlled motion overcomes the actuator generated torque in a seated posture (plotted as a grey line) at the "$\gamma$" zone the standing transition starts with the 1st pulley system engaging at $q_{3std}$ (which varies with the user weight distribution). 
When reaching the standing zone (below $q_3=-90 ^o$) the user has leaning backwards freedom but no leaning forward, as the engaged 1st pulley system reaches a mechanical limit through knees-joint at a set $q_2=90^o$. Around the leaning backwards "$\beta$" zone, the user-initiated motion increases the load making it larger than the provided by the actuator in standing (grey dotted line for the Sitting transition), therefore, a stand-to-sit transition motion starts which engages the 2nd pulley system that ensures support to the torso during sitting, so that, the user hip-joint reaches an angle $q_{3sit}$. 

\subsection{Exoskeleton Setup with Passive Actuator}
The location of a gas-spring actuator was chosen by the system through objective 4, which was fit with multiple providers of commercial actuators. In this case, we chose not to fabricate a custom-made actuator rather, minimize costs and feasibility of applicability of the method by using a discrete number of off-the-shelf actuators.
The significant hysteresis on gas-spring actuators was reported in \cite{Eguchi2018}, thus in Fig. \ref{fig:load_q3} the red-line marks an ideal actuator whereas real torque outputs from data of the gas spring were depicted as a grey line for sit-to-stand and as a dotted grey line for stand-to-sit.
From an ideal actuator equation in (\ref{eq:Fa}) provided by the manufacturer the data of actuation torque showed less than $80 \%$ of effective torque output whereas fitting an equation of measured forces to a third order polynomial showed $99 \%$ confidence as $F_a = (\lambda_0 + \lambda_1  \delta x + \lambda_2 * \Delta x^2)  \eta + D_a \dot{\Delta x}$. Therefore, the optimization process followed the fitted equation with variance in compression and extension forces to achieve an optimal placement.

Using the selected setup we present the results of the model evaluation for the system usability of multiple user-body sizes with variations on height, weight and inertial distribution based on models of inertia distributions in \cite{Leva1996}, implemented through the forward dynamics in (\ref{eq:imp_robot}).

Fig. \ref{fig:COM} shows the COM motion during the full STS, showing only users with full feasible STS in the current configuration. The data exemplify the required small movement by the user of his/her COM of $5$ to $6$ cm on the horizontal axis for initiating the sit-to-stand transition passively assisted. Equally, for stand-to-sit transitions a small variation of 5 to 8 cm is sufficient to control the transition.
Also, the combined human-robot couple variations of the COM with the wheeled based which was an important factor for design of the static stability of the base, and should be consider for the dynamic stability for setting the motion control acceleration constraints.

\subsection{Embodied Standing Mobility Device}
   \begin{figure}[!t]
      \centering
		\subfigure[Lateral view of the mobility device with a virtual user. The trunk support (link 3) in orange, thighs holding (link 2) in black, and the shank base (link 1) in ochre fixed to the mobile base. ]{\includegraphics[width=7.0cm]{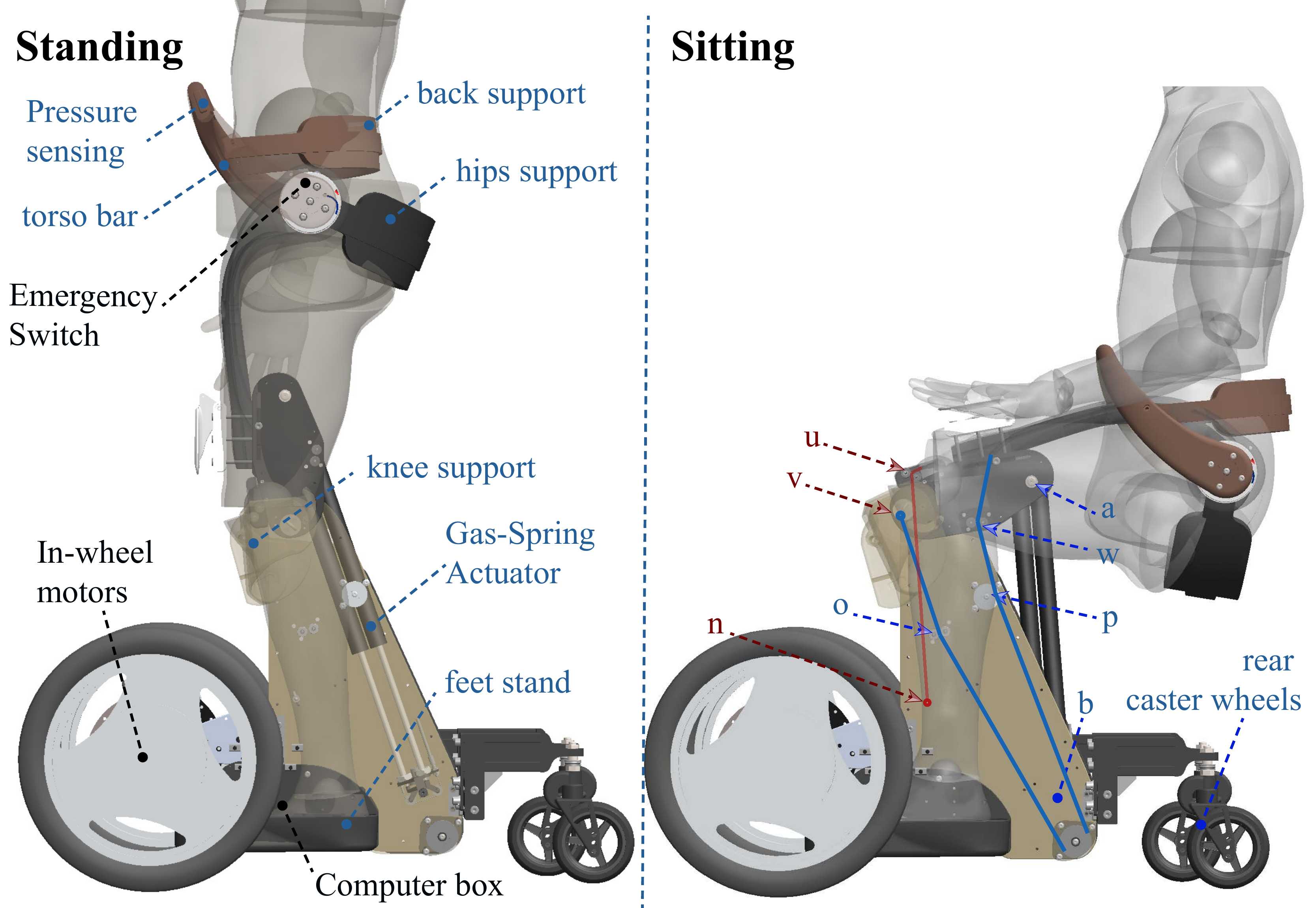}%
    		\label{fig:cad_lat}}
		\hfil
		\subfigure[Detailed view of the parallel cable-driven double-pulley system for transmission of the single actuator torque to both joints.]{\includegraphics[width=7.0cm]{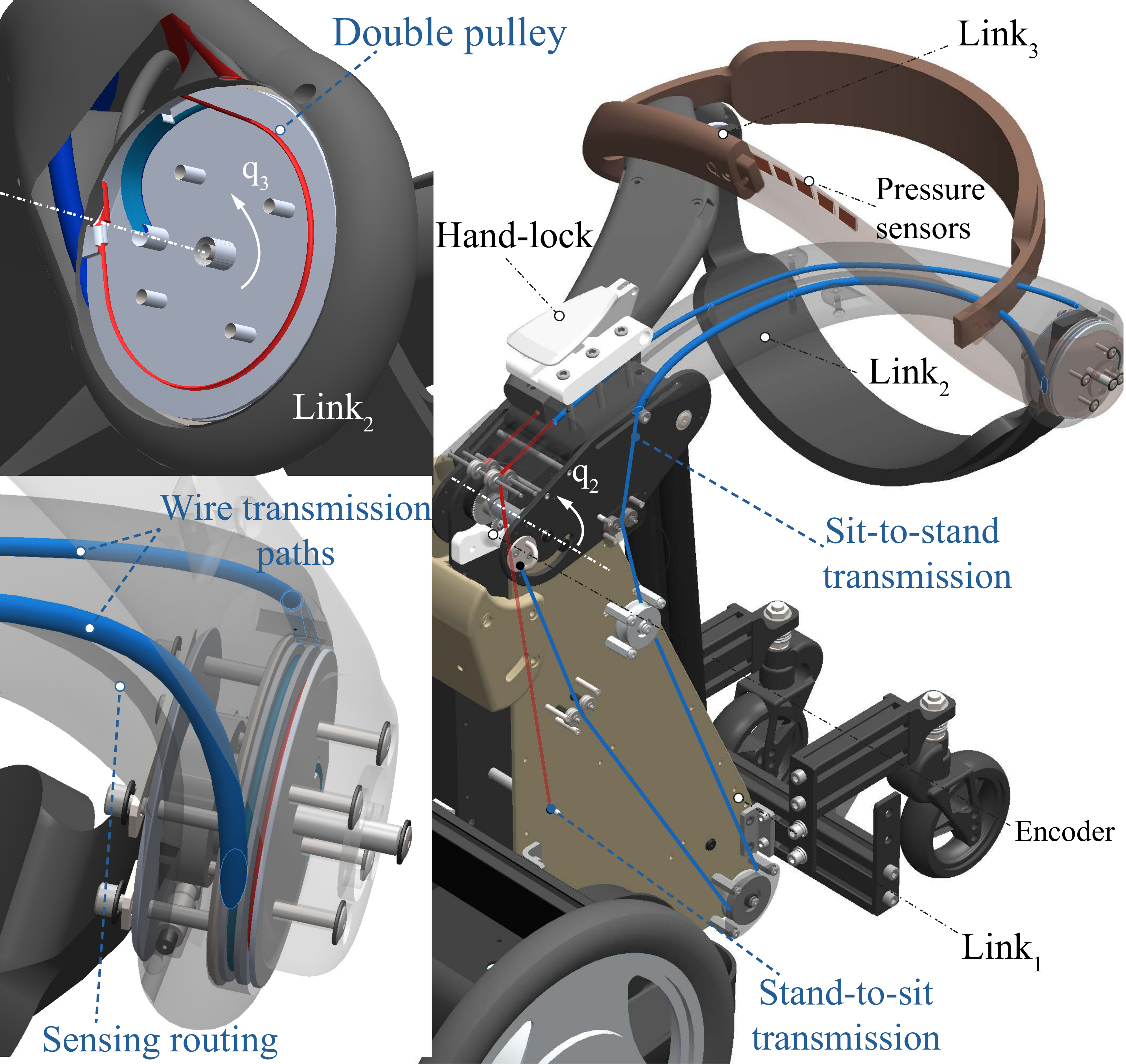}%
		\label{fig:cad_pulley}}
      \caption{Synchronous passive exoskeleton with lower-limbs and torso support with a single passive gas-spring as external energy storage, implemented on a mobile base system for standing mobility.}
      \label{fig:cad_qoloT}
   \end{figure}
The implemented exoskeleton is compact with a 3D printed surfaces on the frontal body (top of the legs and torso) and flexible belts for the rear body part. The overall weight is $6.5 kg$ for links 2 and 3, and $10kg$ for the base link (including three gas-spring actuators).
In order to fit multiple users we developed the current prototype by adjustable mechanism and belts at shank, thighs, buttocks, and back supports, as depicted in Fig.\ref{fig:cad_qoloT} (see the attached multimedia for further details).
The distribution over link 1 and link 2 in Fig.\ref{fig:cad_lat}, shows the assembly result matching the Pareto-optimal result for the wire pulley systems (standing mechanism in blue and sitting in red) and actuator's location. The wire-pulley systems achieved with stainless steel wires ($2 mm$ and $3 mm$) duplicated symmetrically on each side of the exoskeleton.

The double pulley system (Fig. \ref{fig:cad_pulley}) was embedded on a joint $q_3$, and travels within link 2 through routing friction-less tubes.
We achieved a flexible configuration of the engaging angles ($\gamma$ and $\beta$) by changing the length and tension on the wire systems ($P_1$ and $P_2$) through a set of turnbuckles located at link 1 between pulleys ($p-b$) and ($u-n$).
For user safety and ensuring fixed postures, an important component is a locking system through a manual lever located at the thighs, which allows the user to stop the postural transition by blocking the gas-spring power, thus, fixing the exoskeleton at any posture.
As well, we include a end-switch at the standing and sitting posture to ensure the safe locomotion only at standing posture, while any other posture would engage the mechanical brakes.

\subsection{Motion Control System} \label{ss:motion_control}
\begin{figure}[!t]
\centering
\includegraphics[width=7.5cm]{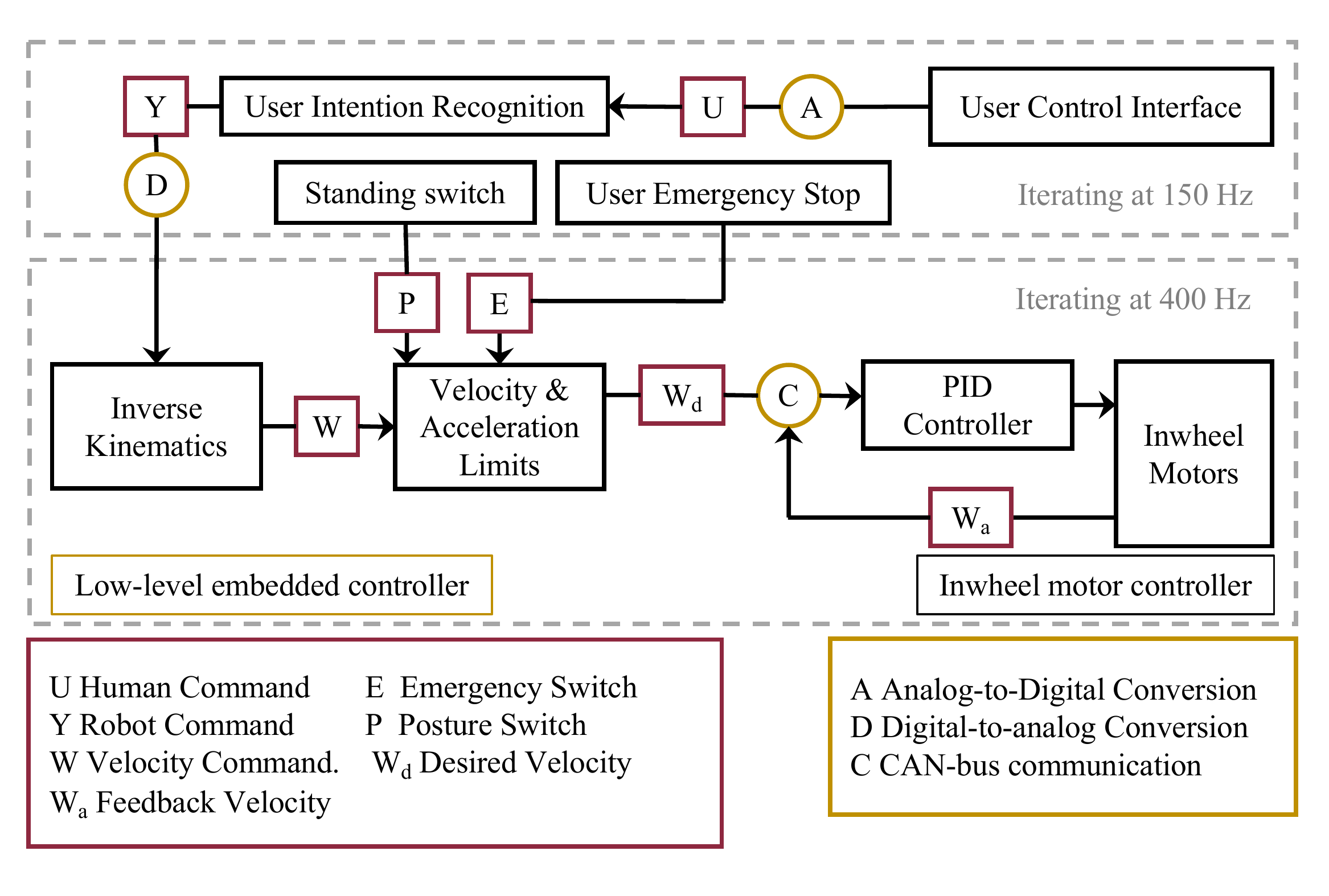}
\caption{Control system architecture, considering the low-level velocity controller for the mobile base of the robot, and the higher-level user intention recognition coupling.}
\label{fig:control_diagram}
\end{figure}
In the current design, we embedded 10 pressure sensors over a soft band around the torso bar for a user's control interface following the description in section \ref{ss:embodied}. The current sensors measure 25$\times$25 mm, without spacing, so that the total covered length is 250 mm (shown in Fig. \ref{fig:cad_pulley}), which fits a wide range of user's waist.
The system was designed to provide high-level intention recognition and results in a desired velocity command (depicted in Fig. \ref{fig:control_diagram}). Thus, the user interface controller was mounted on an embedded computer UP-Squared (AAEON Technology Inc.) using a High-Precision AD/DA Board (Waveshare) for sensors input and communication through analog channels to the actuation low-level control. 
A second low-level controller on an embedded micro-controller circuit handles acceleration and velocity limits and communicating through CANBUS to the wheels velocity controller.
Finally, motion commands are executed on a powered wheeled system using commercial in-wheel motors (JWX-02, YAMAHA Motor Corp., Iwata, Japan) with an internal velocity control loop executing at 400 Hz.

\section{System Evaluation}\label{sec:eval}
\subsection{Sit-to-Stand / Stand-to-Sit Assistance Evaluation}\label{ss:eval_sts}
   	   \begin{figure*}[!t]
      \centering
		\subfigure[Stand-to-sit (above) and sit-to-stand (bottom) transitions using the passive exoskeleton during data recording sessions with volunteer users.]{\includegraphics[width=5.3cm]{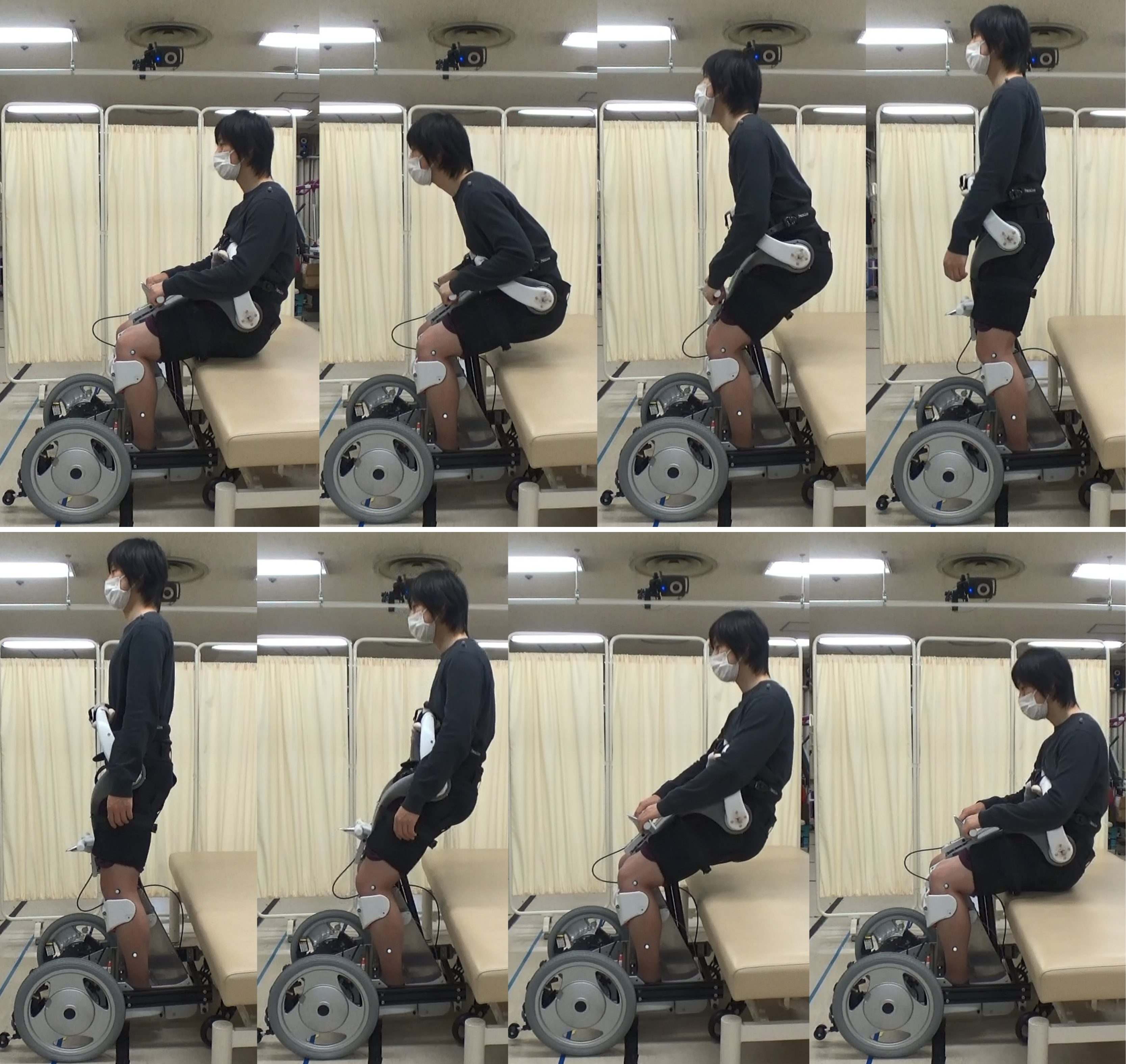}%
    		\label{fig:ss_exp_photo}}
		    \hfil
		\subfigure[Comparison of natural (dotted lines) and exoskeleton assisted (solid lines with SD in shaded areas) STS motions on three users.]{\includegraphics[width=6.0cm]{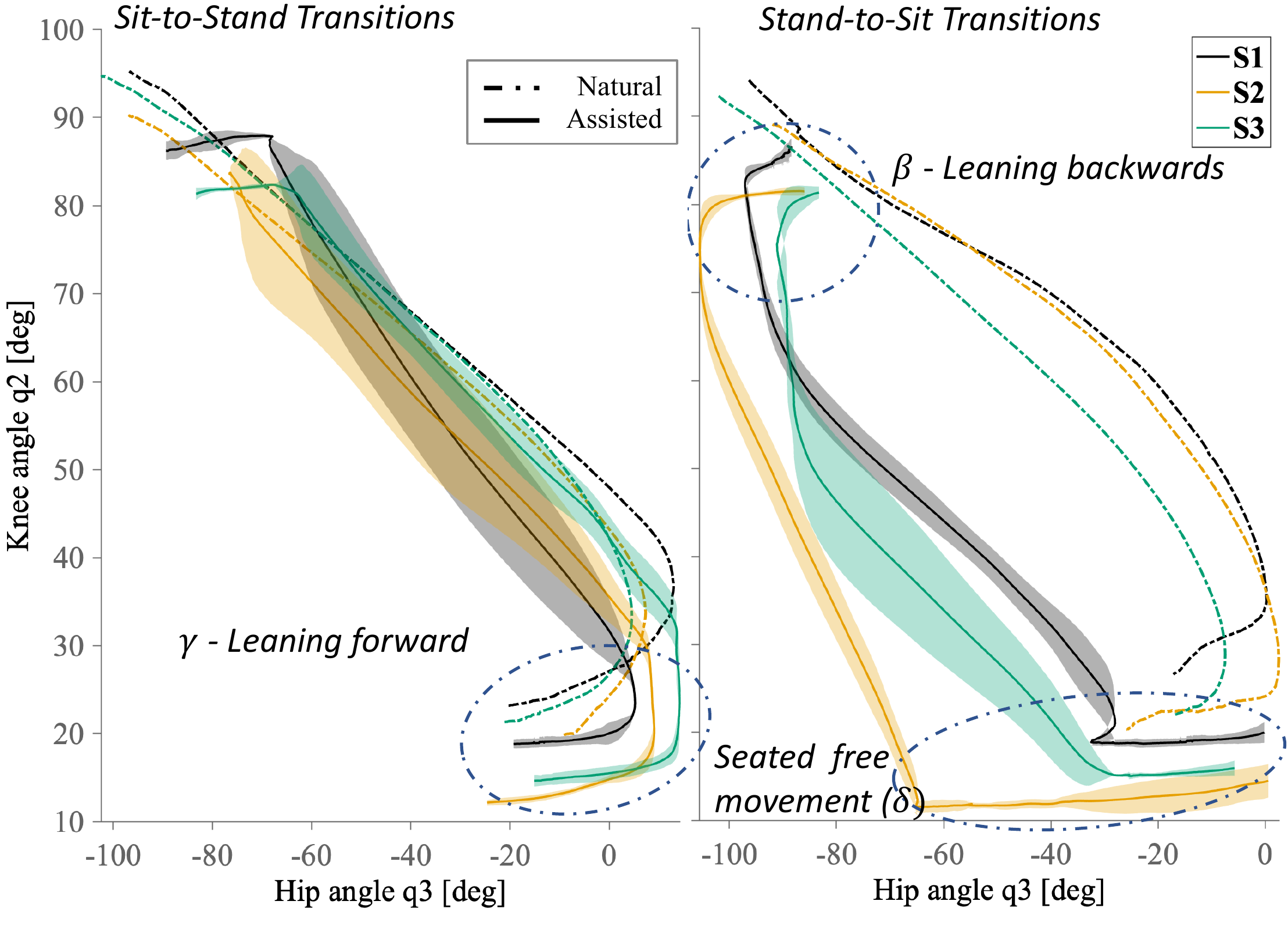}%
		\label{fig:results_STS}}
		    \hfil
		\subfigure[EMG data motion capture data of sit-to-stand and stand-to-sit transition without (left) and with (right) the device from one of the participants. ]{\includegraphics[width=5.7cm]{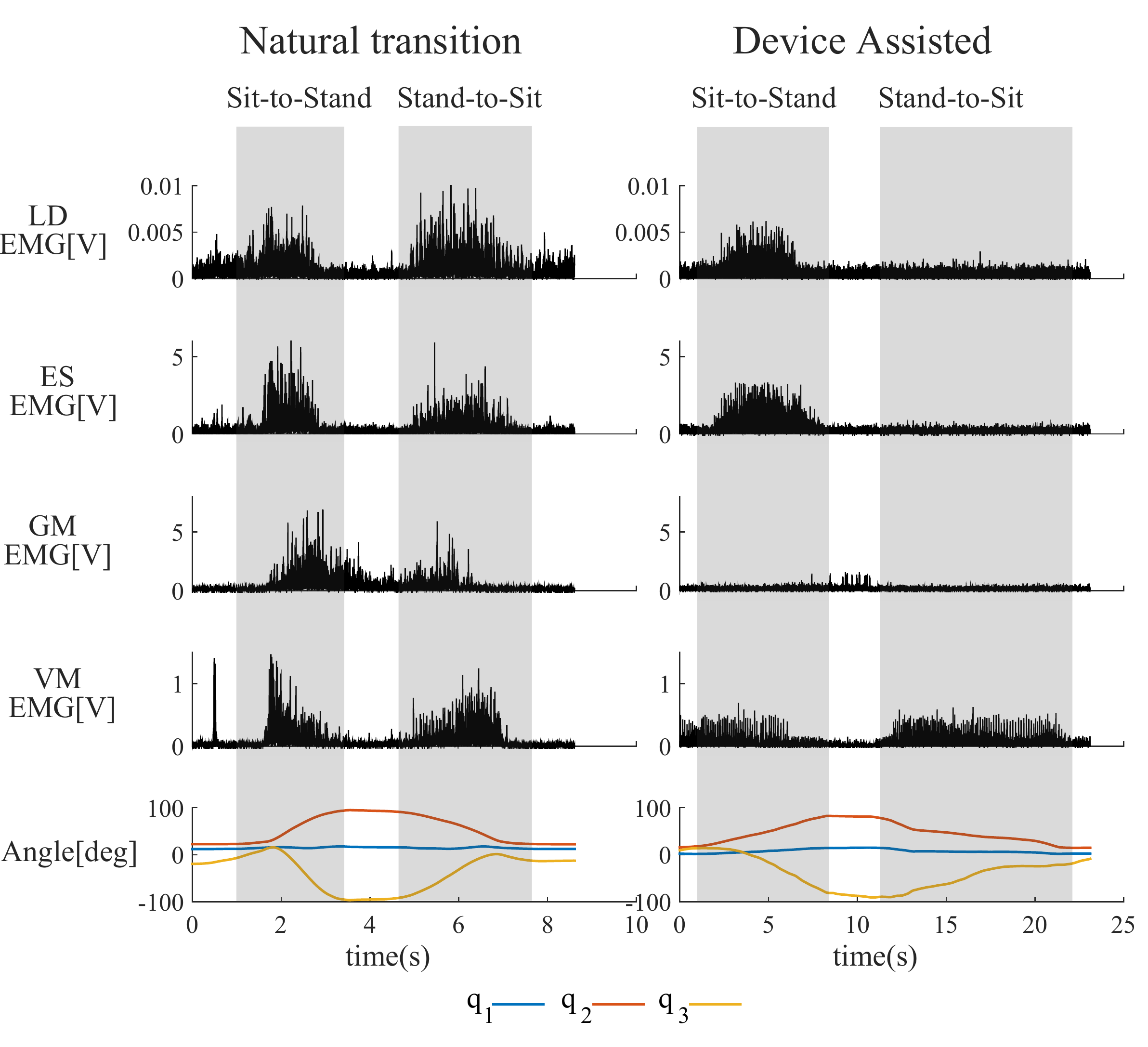}%
    		\label{fig:exp_angle}}
	    \hfil
		\subfigure[Moment load results for a user with 88kg and 1.75 m height. ]{\includegraphics[width=5.5cm]{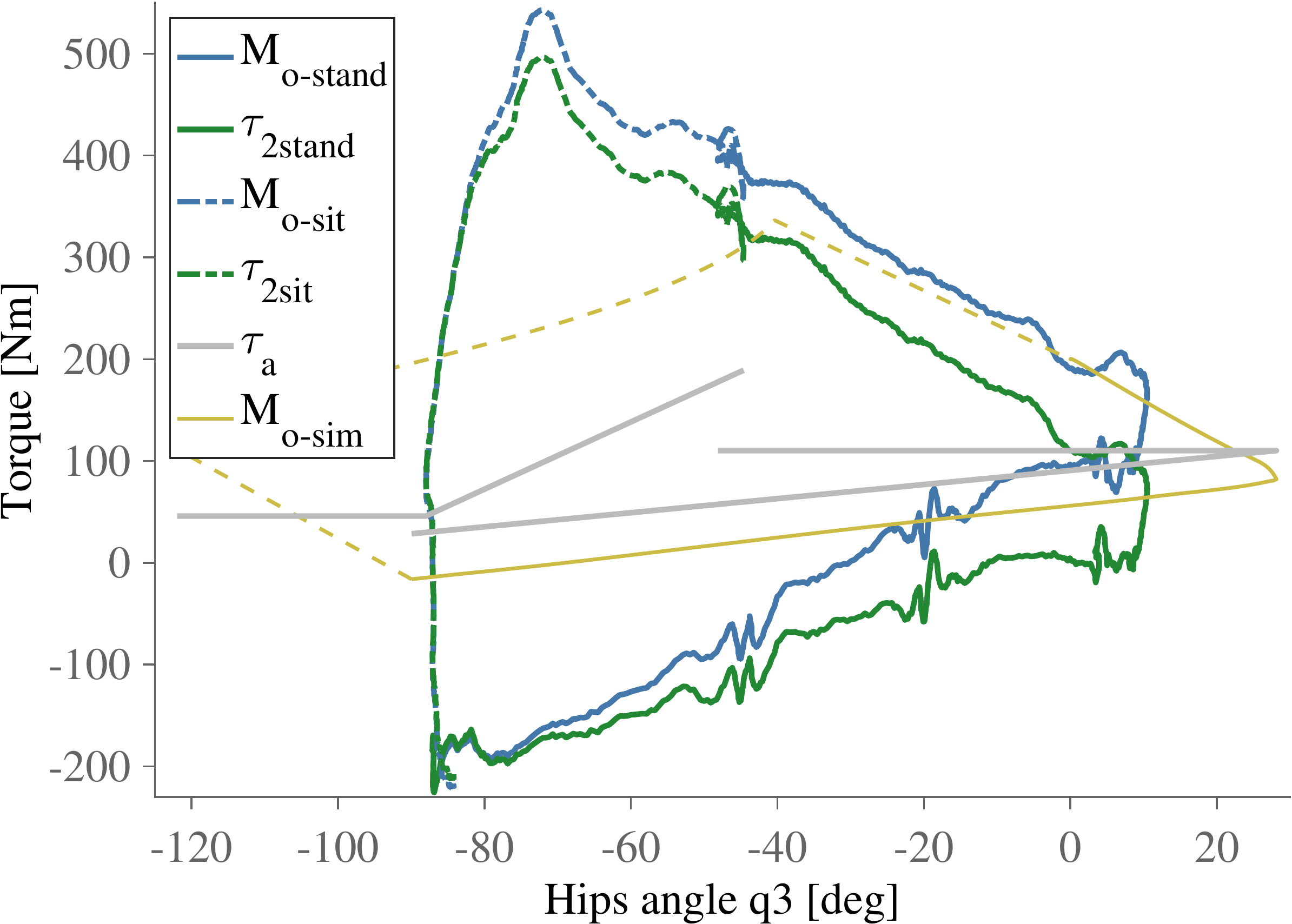}%
		\label{fig:exp_load}}
			\hfil
		\subfigure[Resulting time difference of natural versus assisted postural transitions.] {\includegraphics[width=4.2cm]{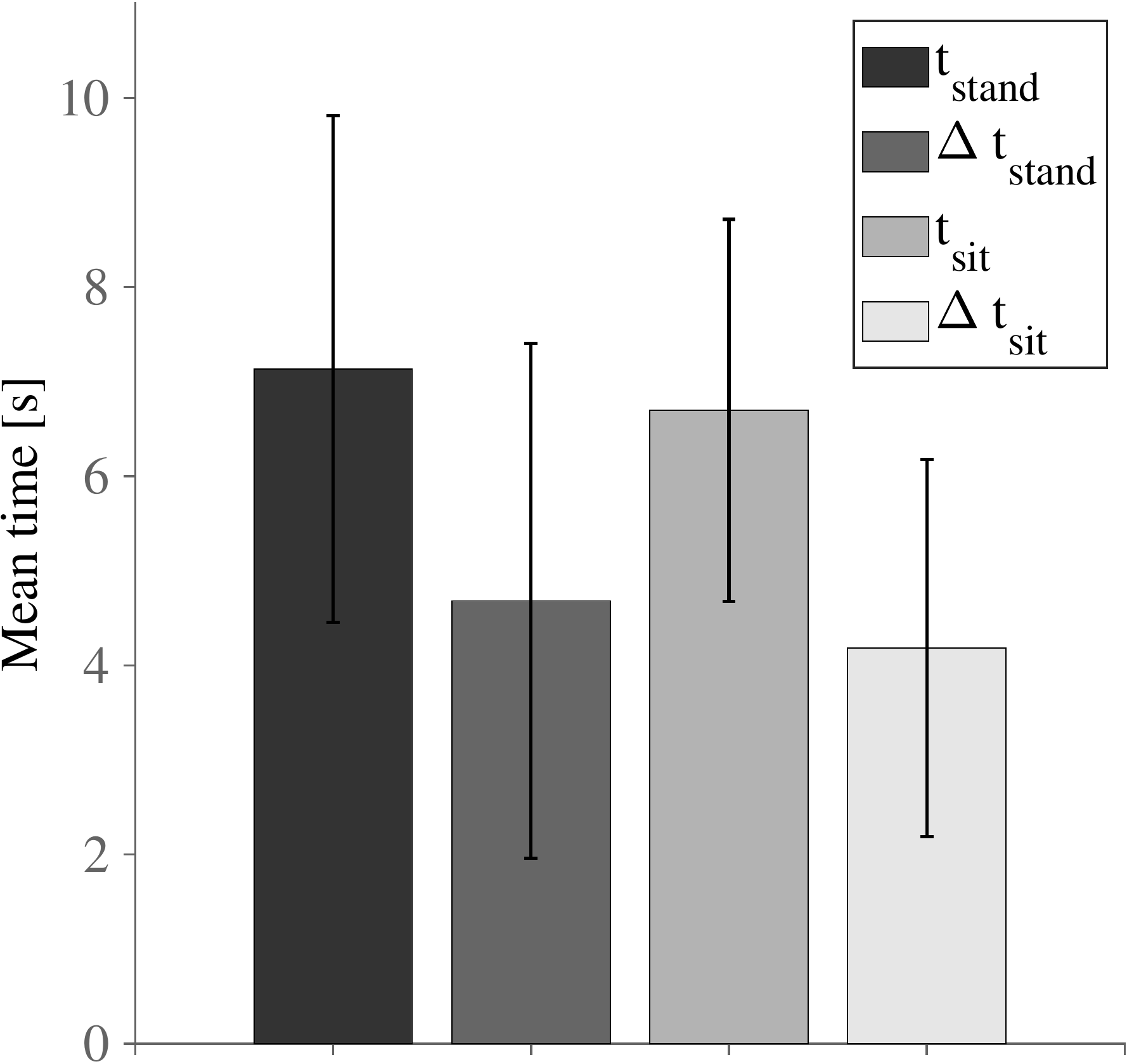}%
		\label{fig:time_stats}}
		    \hfil
		\subfigure[Muscle activity reduction when using the device. Computed from EMG data for standing and sitting transitions. $**$ indicates $p<0.01$.] {\includegraphics[width=4.5cm]{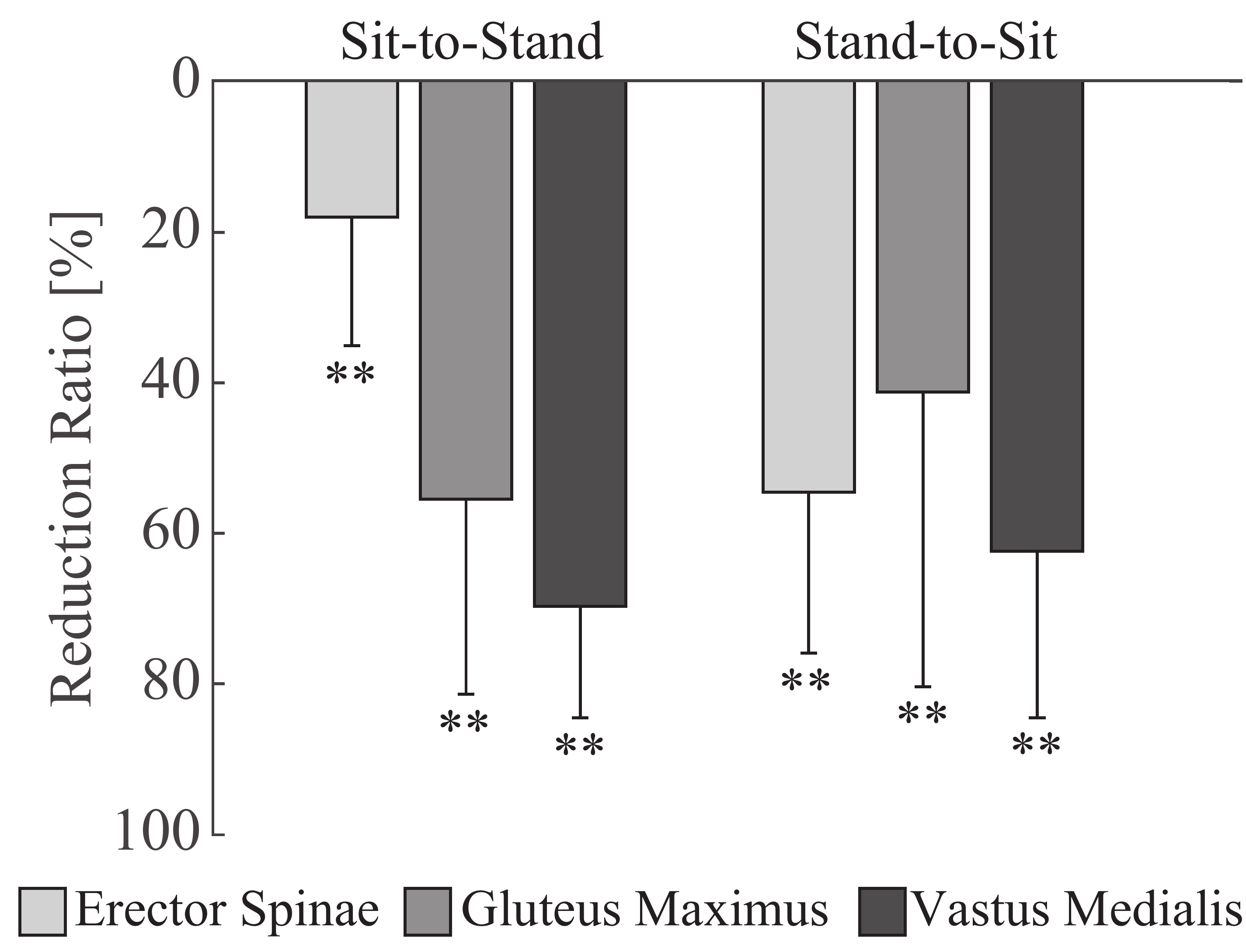}%
		\label{fig:exp_emg}}
		    \hfil
		\subfigure[Maximum leaning forward angle $\gamma$ for standing transitions.]{\includegraphics[width=5.0cm]{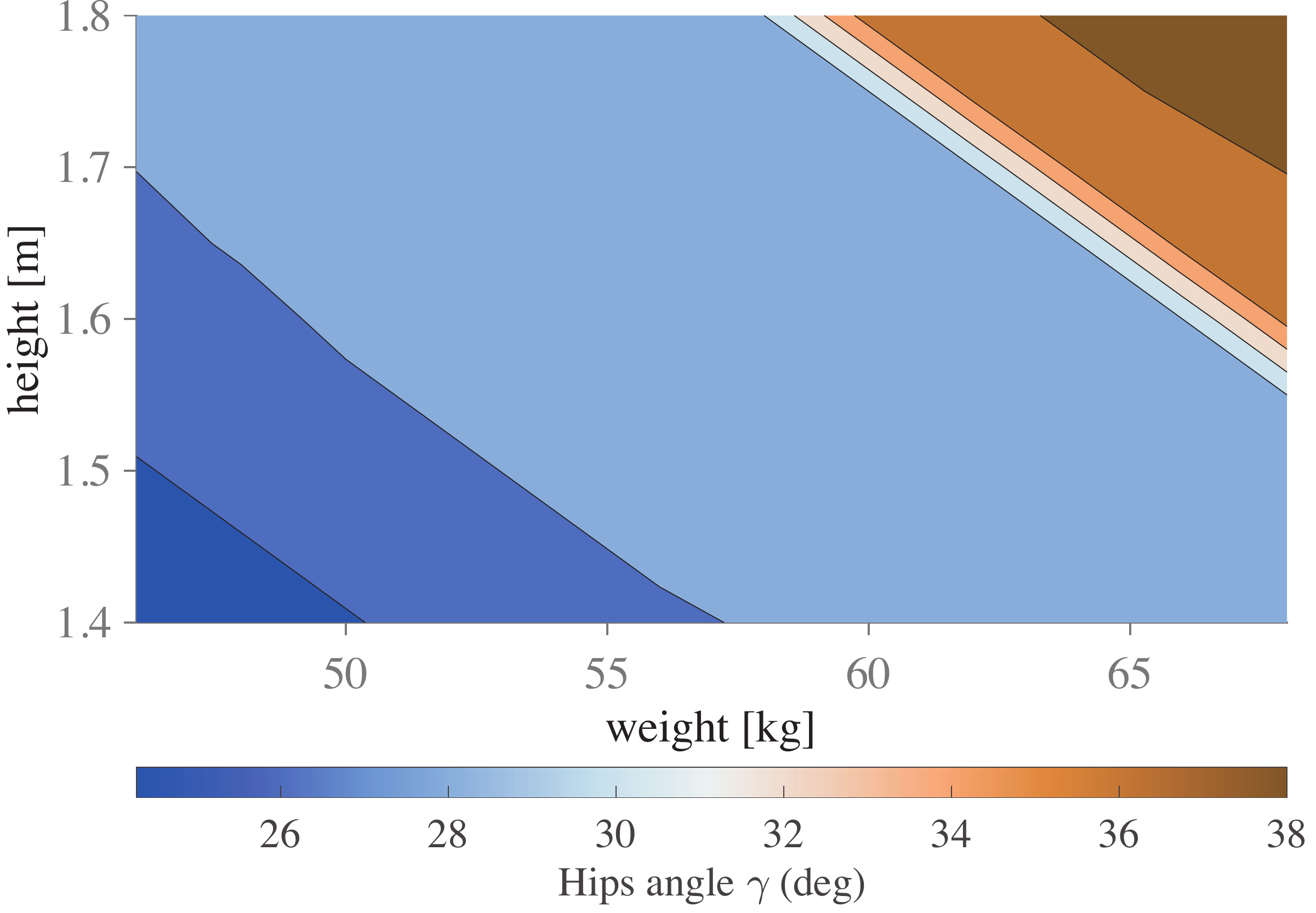}%
		\label{fig:fd_stand_angle}}
		    \hfil
		\subfigure[Maximum leaning backward angle $\beta$ for sitting transitions.]{\includegraphics[width=5.0cm]{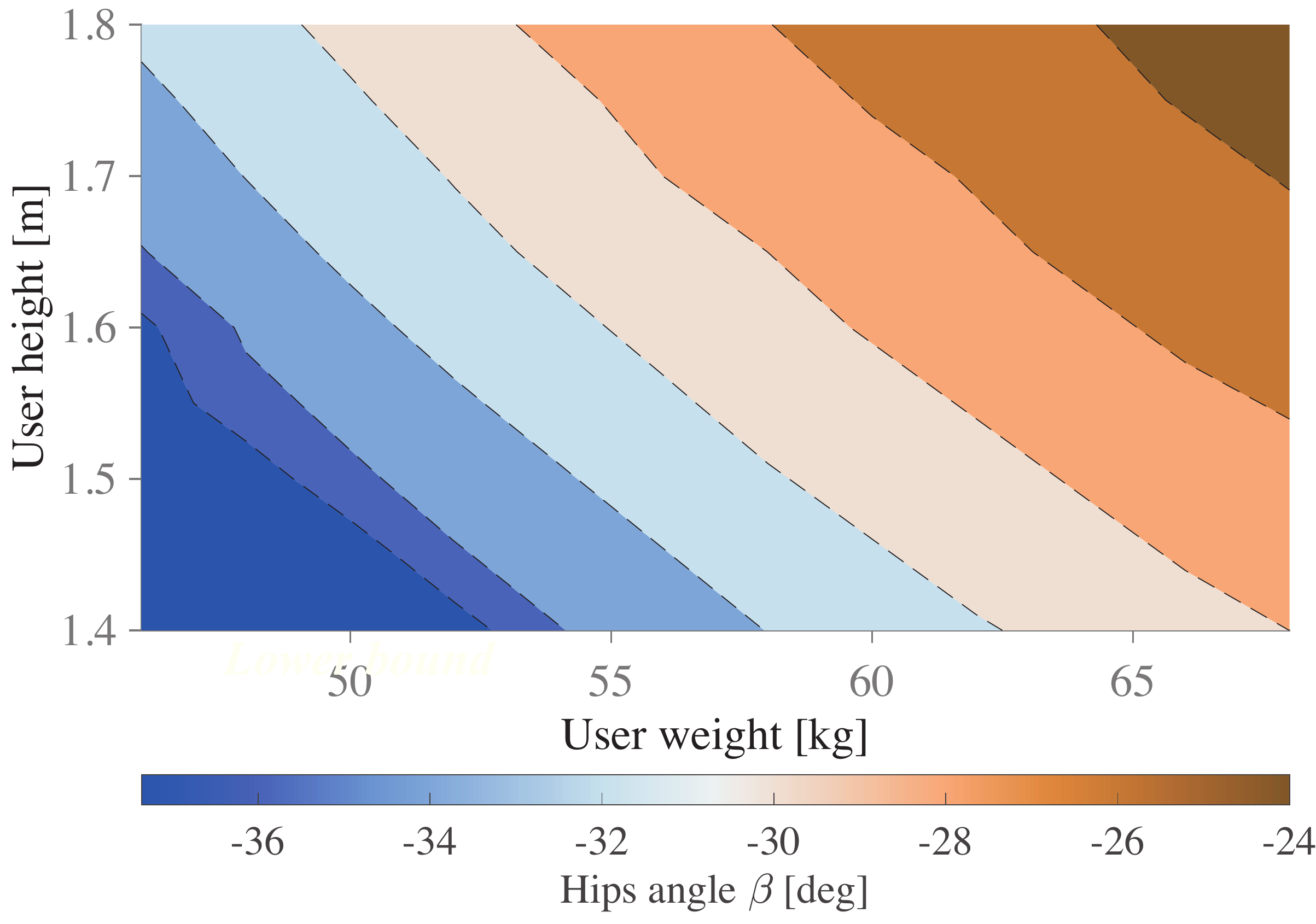}s%
		\label{fig:fd_sit_angle}}
		    \hfil
		\subfigure[Results of assisted motions compared to natural ones.] {\includegraphics[width=4.3cm]{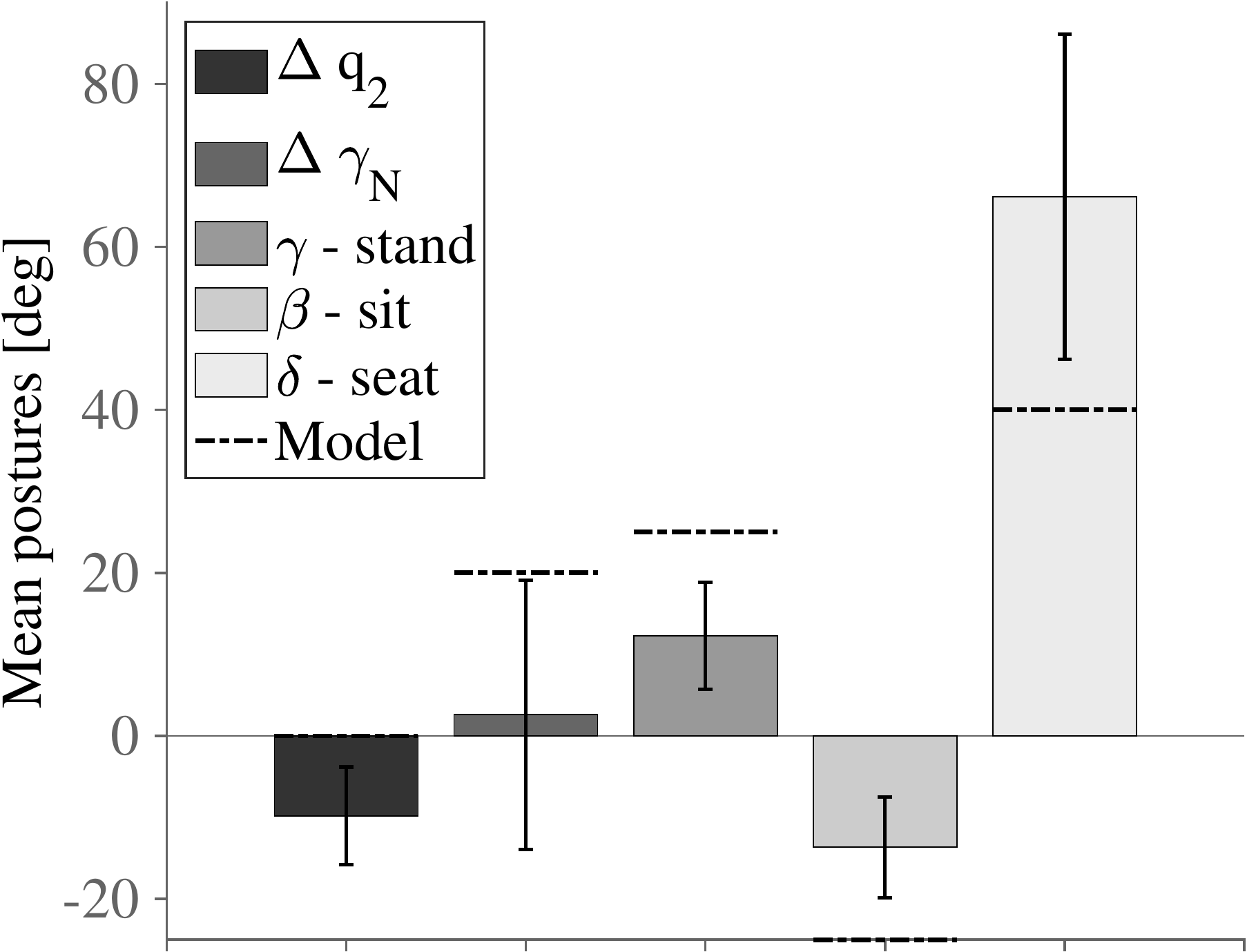}%
		\label{fig:sts_stats}}
		
      \caption{User evaluation study for the proposed synchronous passive exoskeleton with lower-limbs and torso support through a single passive gas-spring as external energy storage, implemented on a mobile base system for standing mobility.}
      \label{fig:experiments}
   \end{figure*}
   	

We conducted experiments with with ten unimpaired male volunteers (age 26.1$\pm$ 4.6 years, 170.3 $\pm$ 8.9 cm, 67.6 $\pm$ 8.1 kg), comparing the necessary muscle groups activity with and without the device, as well as, motion and characteristics of the support. 
The experiments proceeded as follows: each participant performed sit-to-stand and stand-to-sit postural transitions (Fig. \ref{fig:ss_exp_photo}) over three times using the device and then without it (see, Fig. \ref{fig:results_STS}).
The participants were equipped with wireless Electromyography (EMG) sensors (Trigno Lab, Delsys, USA) on the bilateral extensor muscles; latissimus dorsi (LD) and erector spinae (ES) for trunk extension evaluation, gluteus maximus (GM) for hip extension, and vastus medialis (VM) for knee extension.
Markers of a motion capture system (MX System, Vicon Motion Systems, Ltd., U.K.) were attached on the Acromion, Great Trochanter, Lateral Epicondyle of Femur, and Lateral Malleolus to record sagittal motion synchronized with EMG recordings, in order to detect the start and end of transitions (see, Fig. \ref{fig:exp_angle}). 
EMG data were band-pass filtered (40-300Hz), rectified, and evaluated according to maximum value of local integration by a moving window of 100 ms width. 

A paired t-test was adopted to compare between the two conditions. To evaluate the amount of reduction, reduction ratio was computed by $R_r=1-\frac{EMG_{assisted}}{EMG_{without}}$, where $EMG_{assisted}$ and $EMG_{without}$ indicate the EMG with and without the assistance of the device respectively. 

In the assisted sit-to-stand case, maximum activation levels were significantly reduced in average by 18.2\%
for ES, 55.6\% for GM and 68.4\% for VM when compared to that of the unassisted case.
In the assisted case of stand-to-sit, activation levels were reduced in average by 54.7\%, 41.3\% and 62.5\% respectively (Fig. \ref{fig:exp_emg}). With all significant levels taken for ($p < 0.01$).
On the other hand, the activation of LD showed non-significant increase by 99.0$\pm$168.4\% and 33.4$\pm$ 103.2\% in sit-to-stand and stand-to-sit transitions respectively.

Subsequent evaluation of a user ($88kg$, $1.75m$) load to the exoskeleton (see, Fig.\ref{fig:exp_load}) was performed by analysis the acceleration at each joint with additional markers placed at T6, C7, elbow, wrists, and hands, and with measured actuator force output when using 3-gas-springs.
We found a sit-to-stand load normalized moment of 0.026 (well below the design value of 0.86). This is explained because of a significantly high leaning posture in the transition chosen by the subject where leaning forward at $q_4$ and $q_5$ resulted in most of the load quickly converging to the negative plane.
For the stand-to-sit transition, the normalized load was $0.78$, below the expected 0.86. We also observe a different load profile to the modelled motion.
The linearity of the output torque showed an absolute normalized error of 0.098 for the sit-to-stand and 0.039 on the stand-to-sit transition, both below the expected value of 0.45 from simulation data.

\textit{The naturalness of the motion} (Eq. \ref{eq:obj_lin}) was validated among all subjects comparing natural transitions and assisted one (see, Fig. \ref{fig:results_STS}).
The normalized linearity of the coupled motion between joints $q_2-q_3$ on the sit-to-stand transition was $0.11 \pm 0.08$, and  $0.09\pm0.02$ for stand-to-sit, both below the simulated expected value (0.15).
Moreover, comparing to the natural transition of each subject and found and average $\gamma$ of $12.3^{\circ} \pm 6.5^{\circ}$ , which reflects an average $2.6^{\circ} \pm 16.5^{\circ}$ increment from the natural motion. All below the expected values of $25\sim 38^{\circ}$ (see, Fig. \ref{fig:fd_stand_angle}).
The stand-to-sit transitions required an inclination angle $\beta$ of $-13.7^{\circ} \pm 6.1^{\circ}$, ($-6.0^{\circ} \sim -21.7^{\circ}$), all below the expected values of $-24\sim -37^{\circ}$ (see, Fig. \ref{fig:fd_sit_angle}).
The transition times were on average $7.13 \pm 2.7s$ for sit-to-stand, and $6.7 \pm 1.9s$ in stand-to-sit. Which were in average an increment from the natural transitions of $4.6 s$ and $4.1 s$, respectively.

\subsection{Motion Control through Upper-Body Pressure Sensing}\label{ss:eval_navigation}
 \begin{figure}[!t]
    \centering
		\subfigure[Example of pressure sensing from the torso controller with output velocity commands. ]{\includegraphics[width=6.0cm]{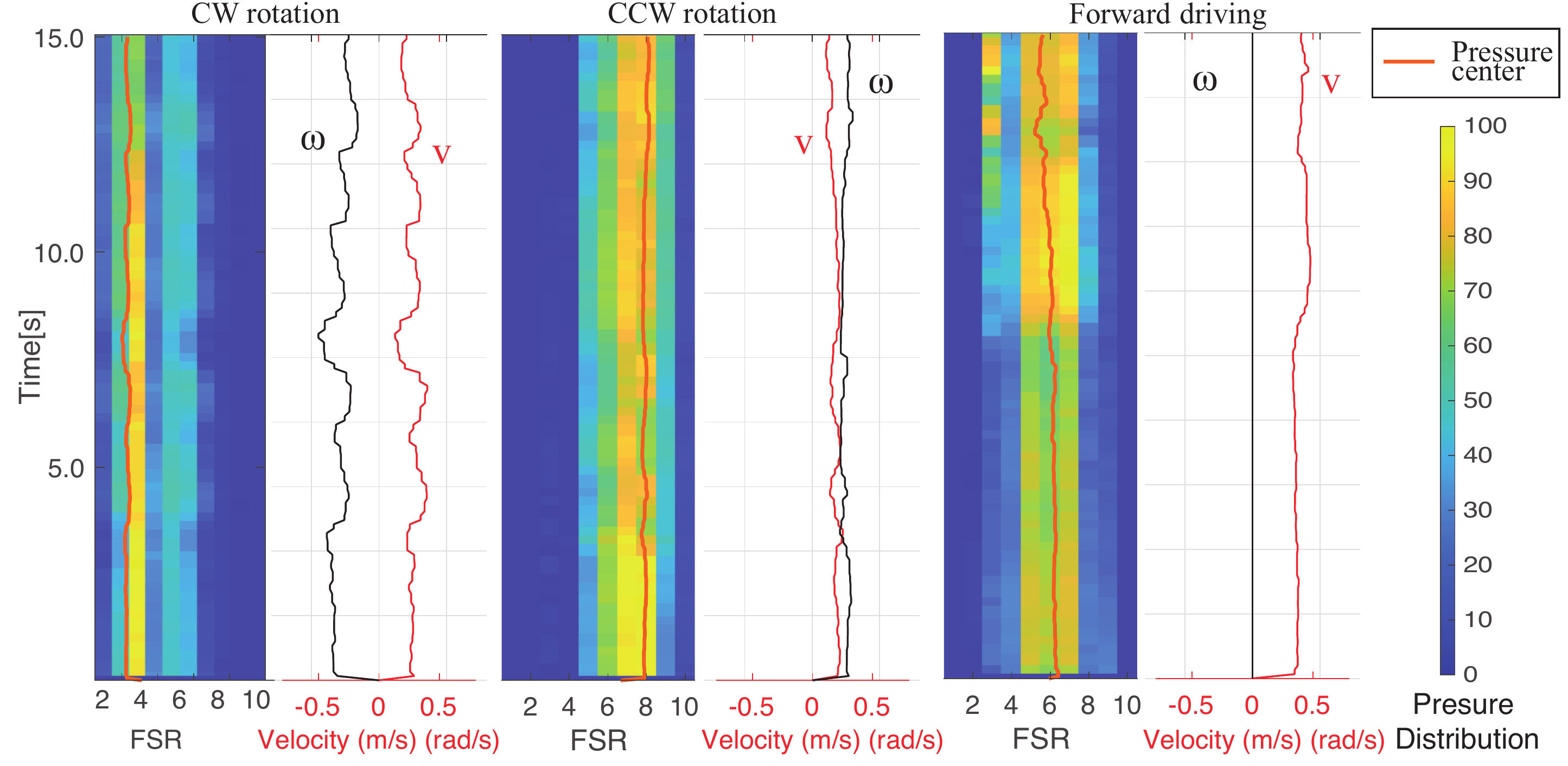}%
    	\label{fig:pressure_distribution}}
		\hfil
		\subfigure[EMG data sample of Erector Spinae, during a left turn (L), right turn (R), and straight (S).]{\includegraphics[width=7.0cm]{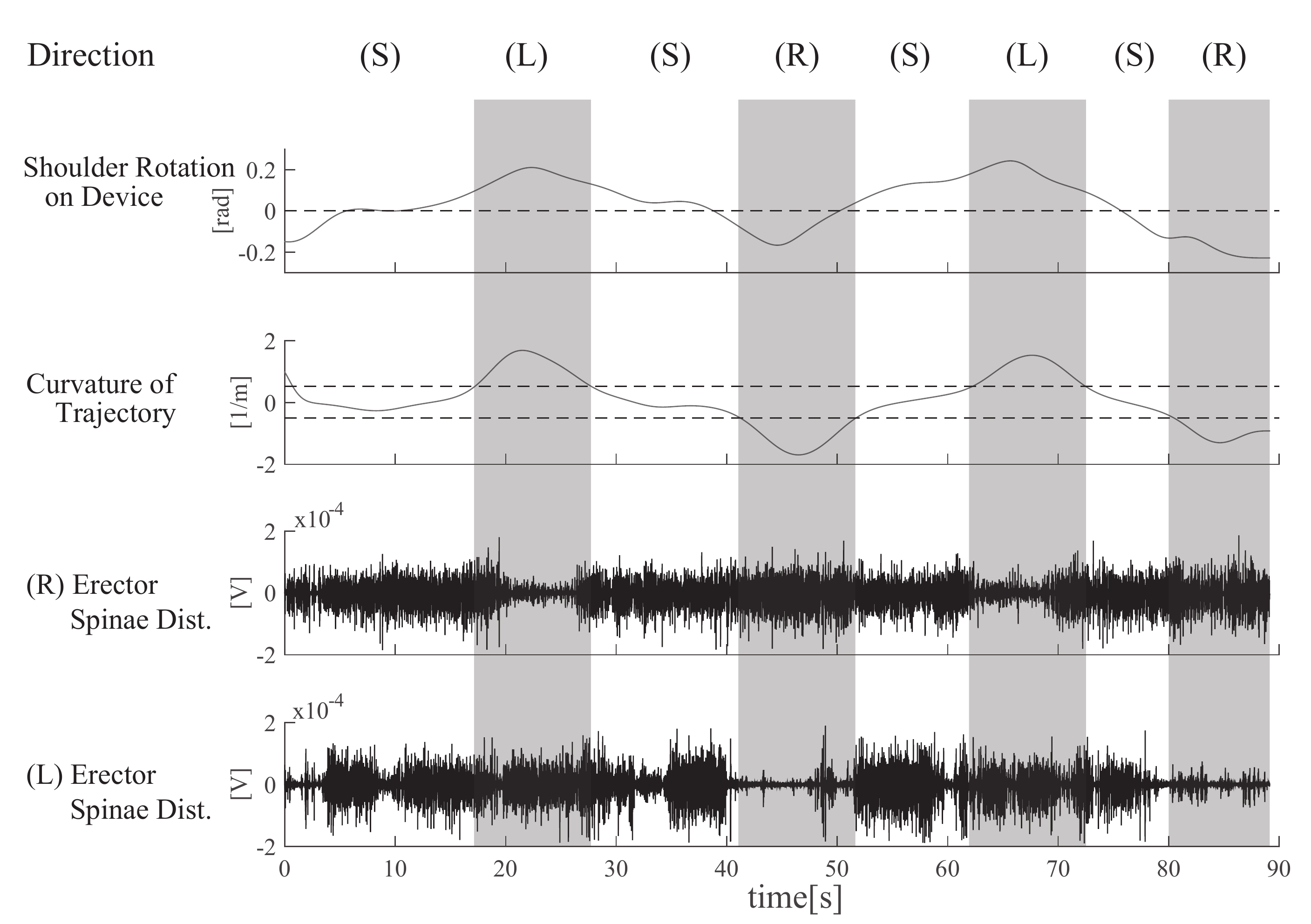}%
		\label{fig:mobility_mocap_emg}}
      \caption{Data example of a user pose to directional control space transformation based on pressure distribution sensing at the lower-torso.}
      \label{fig:mobility_exp}
 \end{figure}
Six unimpaired subjects were recorded performing circular and 8-like navigation motions with the embodied interface of control, as depicted in Fig. \ref{fig:mobility_exp}) with pressure sensing normalized by the sensors range (approximately 0.16$N/cm^2$), and the mapped mobility device velocity response $\xi = [v, \omega]'$.
We found a peak body rotation by comparing shoulder to hip angle on the top plane with a mean of $0.34 rad$ $\pm 0.24 rad$ ($19^{\circ} \pm 13^{\circ}$) among the participants. 
As well, we observed the change in EMG activity at ES and rectus abdominal (RA) muscles during the circuits. Results showed decreased ES activity ($36 \%$ on CCW and $32 \%$ CW) and increased RA activity ($14 \%$) for both rotations compared with straight motions.
These results show the user resulting motion for navigation as proposed through upper-body rotations or stiffening of the abdominal area by means of abdominal contractions and back extensions.

\section{Discussion and Conclusion}\label{s:disscuss}
The proposed light-weighted PMD (36 kg) has a comparable autonomy to that of power wheelchairs (5 to 7 hours) and provides standing locomotion with hands-free navigation through pressure sensing of torso rotations ($\pm 19^{\circ}$). The passive exoskeleton allows natural torso postures for user-controlled STS transitions within short times of $7.3 s $ and $6.7 s $, respectively. Faster than any standing wheelchair ($20 \sim 60 s$).

Compared with our previous designs for ankle-knees support \cite{Eguchi2018} and preliminary torso support for sit-to-stand \cite{PaezGranados2018}, the novel exoskeleton showed a statistically significant reduction of the lower back, hip and knees extensor muscles (ES, GM and VM) during both postural transitions in sitting and standing, compared to natural transitions.
These results suggest the possibility of usage for patients with a significant reduction in the lower back and lower limb muscle control would be able to perform the postural transitions using passive assistance. However, higher levels of mobility impairment might not achieve the device control due to the need for upper back muscles (LD) which showed no reduction.

Results of the dynamic load to the exoskeleton during interaction with a user showed the designed large difference in standing and sitting loads (Fig. \ref{fig:exp_load}) thanks to the asymmetric postural transition while displaying a highly linear load at the knee joint. Although we found a significantly steeper slope on the moment load for the evaluated user  (observed by the crouching of the back and head inclination taken by the user), the transitions were smooth given the damped response on the actuation ($D_a$).

The chosen Pareto optimal configuration showed to perform within the designed parameters in the linearity of the coupled motion of thighs and torso support, with the coupling angles varying per user as expected (see, Fig. \ref{fig:fd_stand_angle}) due to the physiological differences in mass distribution over the body. Nonetheless, the results of the leaning forward ($12.3^{\circ} \pm 6.5^{\circ}$) and backwards angles ($-13.7^{\circ} \pm 6.1^{\circ}$) were lower than simulations for all users, which validates our human models as conservative for the optimization. 
The only posture that exceeded expected values was the sitting angle ($\delta$), with an average of $66.2 \pm 19.9 ^{circ}$, over the expected value of $45^{circ}$. This might be caused by the user's loose attachment to the back support which needs improved fit to the user's height.

Further applications of the proposed exoskeleton and passive assistance methodology through energy balance of the user's motion are envisioned in rehabilitation scenarios through changes in the passive actuation, and further improvement by adapting the system for a wider range of user body types, including women, the elderly and the over-weighted population.
\section*{Acknowledgment}
We thank TESIS graphics for the illustrations and graphical abstract. All user studies were approved by the University of Tsukuba, approval No.2019R300-2.
\ifCLASSOPTIONcaptionsoff
  \newpage
\fi
\bibliographystyle{IEEEtran}
\bibliography{IEEEabrv,TMECH_qolo_T}
\vfill
\begin{IEEEbiography}[{\includegraphics[width=1in,height=1.2in,clip,keepaspectratio]{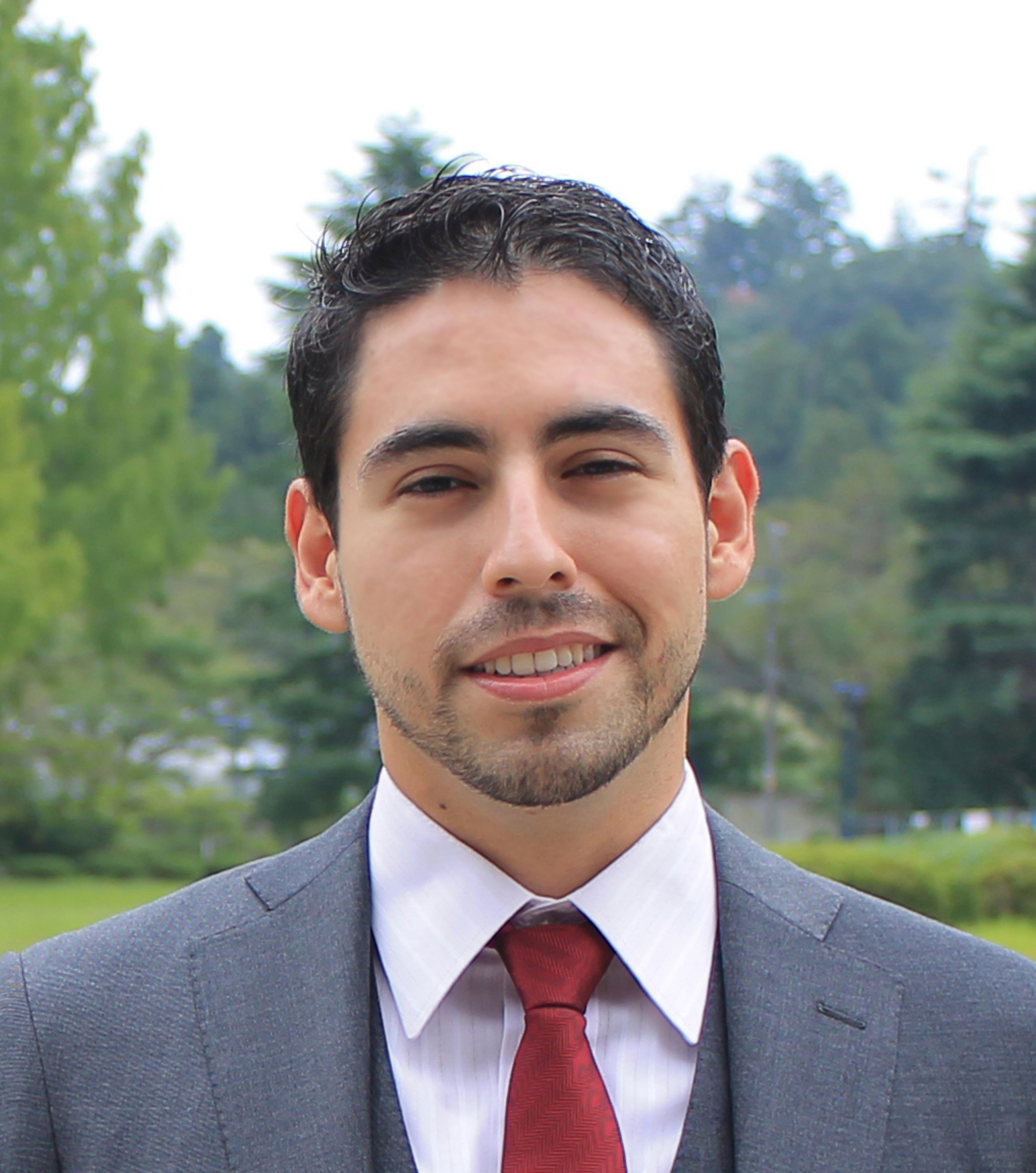}}]{Diego F. Paez-Granados}
received the Ph.D. degree in bioengineering and robotics from Tohoku University, Sendai, Japan, in 2017.
He is currently a Researcher with the Learning Algorithms and Systems Laboratory, EPFL, Switzerland, and a Visiting Researcher with the University of Tsukuba, Tsukuba, Japan. His research interests include physical and cognitive human modeling, compliance in human–robot interaction, shared-control for robot navigation, and soft-robot design and control with human-in-the-loop.
\end{IEEEbiography}

\begin{IEEEbiography}[{\includegraphics[width=1in,height=1.2in,clip,keepaspectratio]{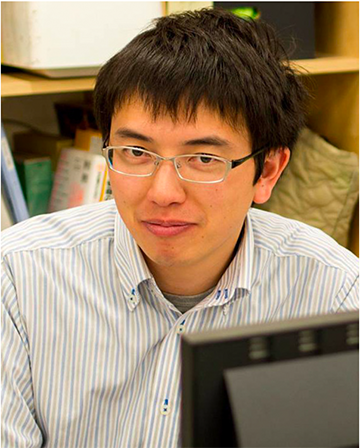}}]{Hideki Kadone}
received his Ph.D. degree in information science and technology from the University of Tokyo, Japan, in 2008. He is an associate professor in Faculty of Medicine, and Center for Cybernics Research, at University of Tsukuba. His research interests include clinical motion measurement and analysis and development of wearable assistive devices to compensate for, support or improve motor function of patients after neurological or mechanical disorders.
\end{IEEEbiography}

\begin{IEEEbiography}[{\includegraphics[width=1in,height=1.2in,clip,keepaspectratio]{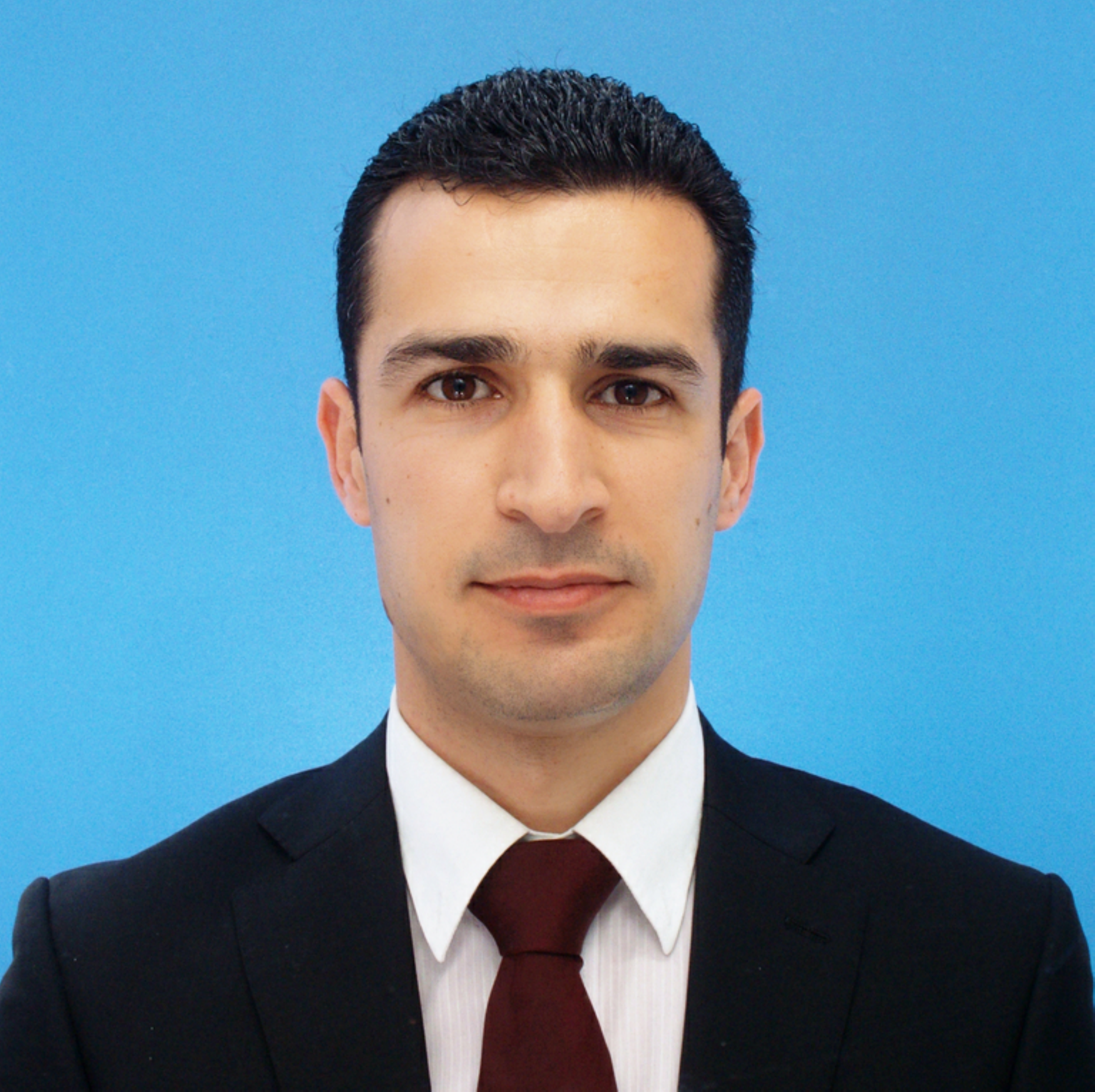}}]{Modar Hassan}
received the BS degree in Engineering in 2009 from Tishreen University Syria. He received ME and PhD in Engineering in 2013, 2016, and MS in Medical Sciences in 2016 from the University of Tsukuba, Japan. From 2010 to 2011 he was a lecturer of Mechatronics at Tishreen University, Syria. He is currently a assistant professor at the Faculty of Engineering, Information and Systems, University of Tsukuba, Japan. His research interests include assistive robotics, orthotics, biomechanics, and motor control. He is a member of the IEEE.
\end{IEEEbiography}

  \begin{IEEEbiography}[{\includegraphics[width=1in,height=1.2in,clip,keepaspectratio]{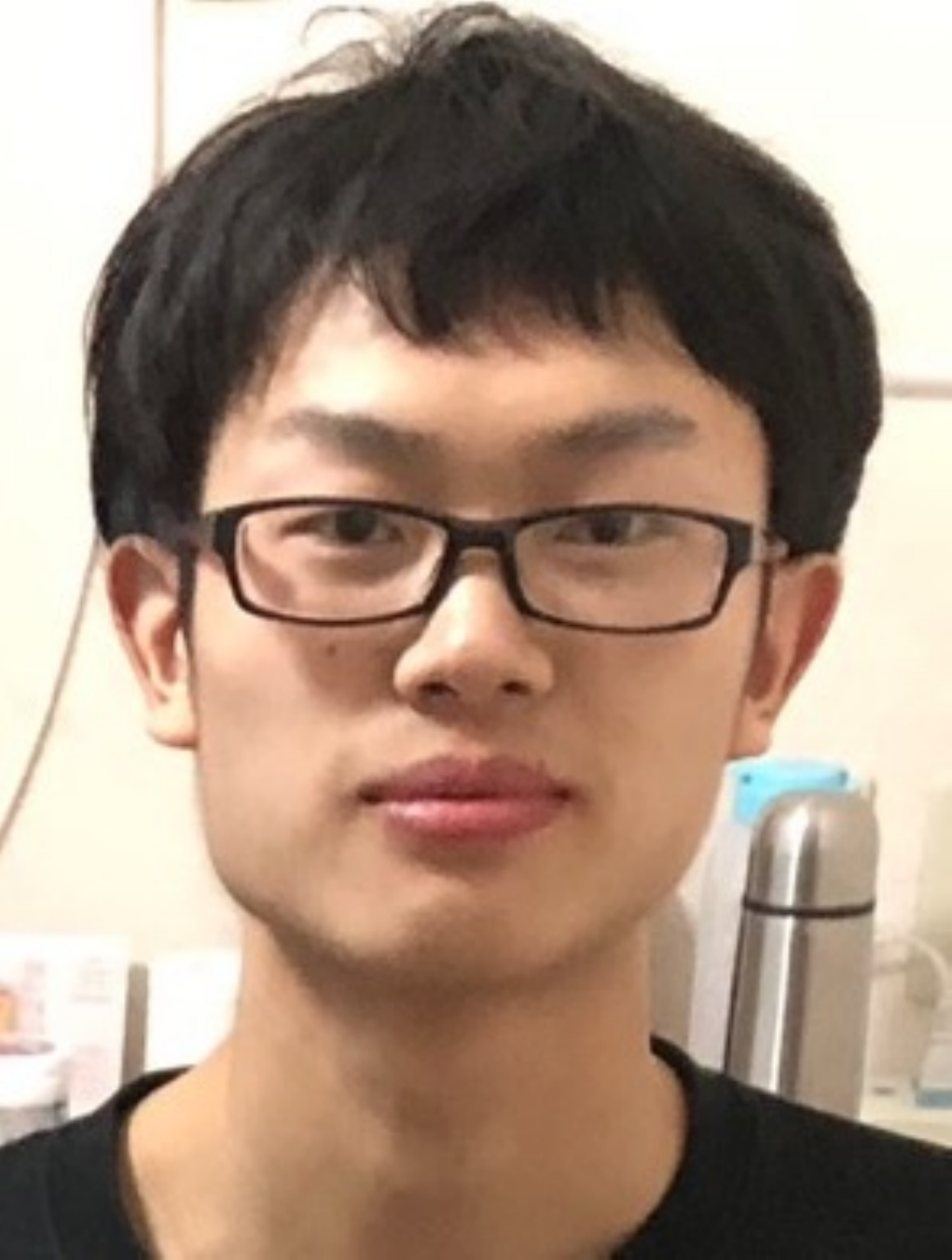}}]{Yang Chen}
received the B.Eng. in mechanical engineering from Jilin University, China, in 2017 and his master degree in human informatics  from the University of Tsukuba, Japan, in 2020 where he is currently working toward the Ph.D. degree. His research interests include human assistive device, mobile robot and autonomous control. 
  \end{IEEEbiography}

\begin{IEEEbiography}[{\includegraphics[width=1in,height=1.2in,clip,keepaspectratio]{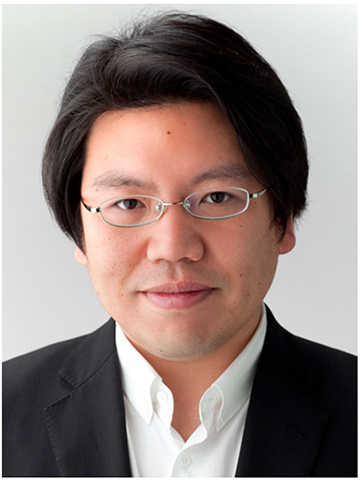}}]{Kenji Suzuki}
received his PhD degrees in pure and applied physics from Waseda University, Japan, in 2003. He is currently a full professor in the Center for Cybernics Research, and principal investigator of Artificial Intelligence Laboratory, Faculty of Engineering, University of Tsukuba, Japan. He was also a visiting researcher at the Laboratory of Physiology of Perception and Action, College de France in Paris, and the Laboratory of Musical Information, University of Genoa, Italy. His research interests include augmented human technology, assistive and social robotics, humanoid robotics social playware, and affective computing.
\end{IEEEbiography}


\end{document}